\documentclass{article}

\usepackage[final]{neurips_2024}
\usepackage{abbrv}

\usepackage[utf8]{inputenc} 
\usepackage[T1]{fontenc}    
\usepackage{hyperref}       
\usepackage{url}            
\usepackage{booktabs,arydshln} 
\usepackage{amsfonts}       
\usepackage{nicefrac}       
\usepackage{microtype}      
\usepackage{xcolor}         
\usepackage{graphicx}
\usepackage{graphics}
\usepackage{placeins}

\usepackage{natbib}
\setcitestyle{square,numbers,sort,comma}
\usepackage{multirow,tableutils}
\usepackage{sidecap,array,tabularx,caption}
\usepackage[nointegrals]{wasysym}
\usepackage{tikz, tikzfigs}

\usepackage{caption}
\DeclareCaptionType{eqcap}[][List of equations]
\usepackage{amsmath}
\usepackage{amssymb}
\usepackage{mathtools}
\usepackage{physics}
\usepackage{algorithm}
\usepackage{algpseudocode}
\usepackage{siunitx}
\usepackage{acronym, acronymdefs}

\usepackage{hyperref}
\usepackage[capitalize,noabbrev]{cleveref}

\usepackage{orcidlink}
\hypersetup{
  colorlinks=true,
  citecolor=[rgb]{0,0.55,0.27},
  urlcolor=[rgb]{0,0.55,0.27},
  linkcolor=[rgb]{0,0.55,0.27}
}

\title{Characterizing Model Robustness via Natural Input Gradients}

\author{%
  Adrián Rodríguez-Muñoz \\
  MIT CSAIL\\
  \texttt{adrianrm@mit.edu} \\
  \And
  Tongzhou Wang \\
  MIT CSAIL \\
  \texttt{tongzhou@mit.edu} \\
  \AND
  Antonio Torralba \\
  MIT CSAIL \\
  \texttt{torralba@mit.edu} \\
}

\begin{document}

\maketitle

\begin{abstract}
Adversarially robust models are locally smooth around each data sample so that small perturbations cannot drastically change model outputs. In modern systems, such smoothness is usually obtained via \emph{Adversarial Training}, which explicitly enforces models to perform well on perturbed examples. In this work, we show the surprising effectiveness of instead regularizing the gradient \wrt model inputs \emph{on natural examples only}. Penalizing input \emph{Gradient Norm} is commonly believed to be a much inferior approach. Our analyses identify that the performance of \emph{Gradient Norm} regularization critically depends on the smoothness of activation functions, and are in fact extremely effective on modern vision transformers that adopt smooth activations over piecewise linear ones (\eg, ReLU), contrary to prior belief. On ImageNet-1k, \emph{Gradient Norm} training achieves $> 90\%$ the performance of state-of-the-art PGD-3 \emph{Adversarial Training} (52\% vs.~56\%), while using only $60\%$ computation cost of the state-of-the-art without complex adversarial optimization. Our analyses also highlight the relationship between model robustness and properties of natural input gradients, such as asymmetric sample and channel statistics.
Surprisingly, we find model robustness can be significantly improved by simply regularizing its gradients to concentrate on image edges without explicit conditioning on the gradient norm. 
\end{abstract}

\section{Introduction}
\label{sec:intro}

Deep neural networks have become the gold standard for computer vision tasks due to their strong empirical performances. However, they are also extremely brittle. Adversarial examples \cite{szegedy_intriguing_2014} are small manipulations to input images that can cause highly-performant models to fail catastrophically. For example, the accuracy of an ImageNet deep classifier drops from 84\% to 0\% under such attacks, even though the perturbed images look identical to humans.

To safely deploy these models in critical tasks such as medicine or autonomous vehicles, extensive research has been devoted to making robust models that are invariant to these small perturbations. The current foremost paradigm for obtaining robust models in practice has been \emph{Adversarial Training} \cite{madry_towards_2018, athalye_obfuscated_2018}, which trains models in a minimax fashion by optimizing classification losses over the attack-perturbed images. While this approach is effective (yielding 60\% robust accuracy under attack), it is also extremely computationally expensive, taking $3.92 \times$ wallclock time per training iteration compared to normal training (see \cref{tab:grad-norm-computational-cost}). Therefore, it is important to seek properties of robust models that can be optimized much more efficiently.

\begin{figure}[t!]
    \centering
    \includegraphics[width=\linewidth]{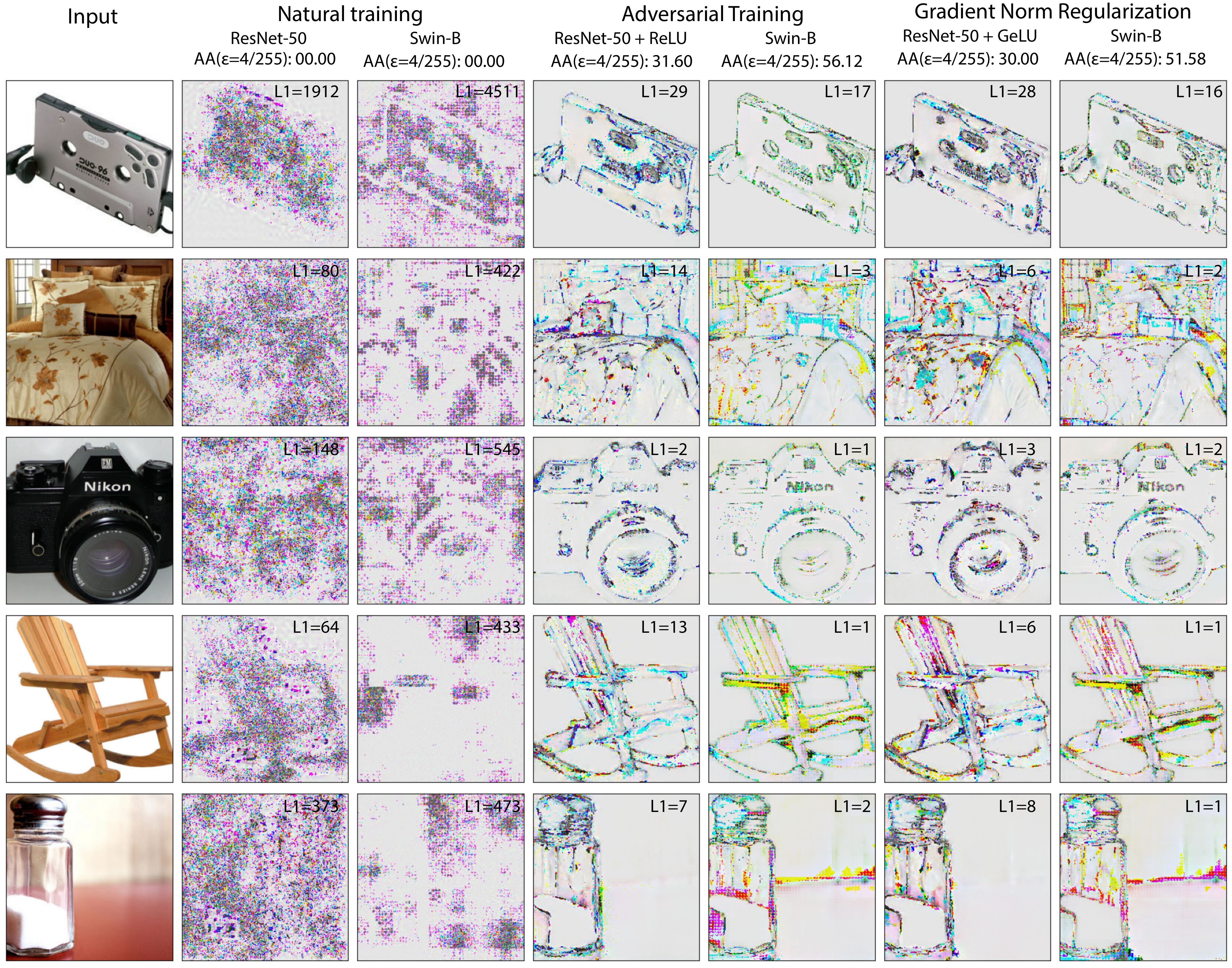}
    \caption{Comparison of loss-input gradients of non-robust and robust models across architectures for a set of images. Non-robust models are taken from the open-source repository \texttt{timm} \cite{wightman_pytorch_2019}. Adversarial training is from the work of Liu \etal \cite{liu_comprehensive_2023}. Gradient norm regularization optimizes \cref{eq:gradnorm-objective}. As can be seen, a model can be easily identified as vulnerable or robust simply by looking at clean input gradients. Gradients of robust models (adversarial training and gradient norm regularization) highly resemble the input images, and look visually similar to each other to the human eye. By contrast, gradients of vulnerable models are noise-like, bearing apparently little resemblance to each other or the input images. Numerically, the norm of the input gradient (top right for each gradient) is also highly discriminative of vulnerability or robustness. Gradients normalized to [0, 1] then shifted by 0.4 for visualization purposes only.} 
    \label{fig:gradient-comparison}
\end{figure}

In this work, we aim to characterize model robustness via \emph{loss-input gradients} \citep{simon-gabriel_first-order_2019, ross_improving_2017}:
\begin{equation}
    \nabla_x \mathcal{L} \coloneqq \underbrace{\nabla_x\mathcal{L}_{\text{CE}}(f_{\theta}(x), y)}_{\mathclap{\textsf{loss-input gradient of model $f_\theta$ on example $x$ with groundtruth class $y$}}},
    \label{eq:loss-input-gradient}
\end{equation} 
where $\mathcal{L}_{\text{CE}}$ is the cross entropy loss.  The quantity $\nabla_x \mathcal{L}$ is related to the first-order Taylor expansion of model loss. It is known that a smaller \emph{Gradient Norm} $\norm{\nabla_x \mathcal{L}}$ is correlated to model smoothness and thus robustness \citep{simon-gabriel_first-order_2019, ross_improving_2017}. 
\cref{fig:gradient-comparison} highlights the substantial visual differences between the loss-input gradients of vulnerable and robust models of the same architecture. Such differences reveal the regions of the image that the model uses to make predictions. Notably, the gradients for robust models are more ``image-like''. 
In robust models, the gradients are generally focused on the edges of input images, and in fact, have much smaller magnitudes (\Cref{tab:evidence-low-gradient-norms}). In this paper, we analyze if these properties contribute to robustness, or are simply irrelevant byproducts.

Previous research \citep{seck_l_2019} showed that simply regularizing $\norm{\nabla_x \mathcal{L}}$ only yielded limited robustness compared to \emph{Adversarial Training}, the current state of the art. We revisit this regularized training objective, and find it very effective on model architectures that use smooth activation functions, including modern vision transformers \citep{liu_swin_2021}. In contrast to prior beliefs, our results on ImageNet \cite{deng_imagenet_2009} show that more than 90\% of robustness (compared to the state of the art) can be obtained by simply regularizing input gradients using 63\% of the computational cost (see \cref{tab:grad-norm-computational-cost}) and no inner maximization. Beyond regularizing magnitudes, we also obtain non-trivial robustness gains by enforcing gradients to concentrate on image edges in \cref{sec:edge-regularization}. Our work is the first to show the importance of input gradients $\nabla_x \mathcal{L}$ in both understanding and improving the robustness of modern neural network architectures. 

In summary, our contributions are as follows: \begin{enumerate}
    \item We are the first to achieve near state-of-the-art robustness on ImageNet by only regularizing the norm of \emph{natural} input gradients (\cref{sec:regularizing-for-small-gradient-norms}).
    \item We demonstrate that the effectiveness of gradient norm regularization critically depends on the smoothness of activation functions (\cref{sec:effect-of-smoothness}). 
    \item We show that significant robustness appears when input gradients are more perceptually ``image-like'' and enforced to concentrate on image edges without using any other regularization (\cref{sec:edge-regularization}). 
\end{enumerate}

\section{Related Works}
\label{sec:related-work}

\paragraph{Adversarial examples.}
Szegedy \etal \citep{szegedy_intriguing_2014} first identified the existence of \emph{adversarial examples}, small perturbations imperceptible to humans but that completely fool networks. Since then, extensive researches have been conducted on the subject of adversarial examples, both on defending against such attacks \cite{kurakin_adversarial_2017, tramer_ensemble_2018, madry_towards_2018, guo_countering_2018, xie_mitigating_2018, wong_provable_2018, liao_defense_2018, cohen_certified_2019, zhang_adversarial_2019, pang_bag_2021}, as well as stronger attacks \cite{goodfellow_explaining_2015, kurakin_adversarial_2016, carlini_evaluating_2019,dong_boosting_2018, xie_improving_2019, croce_reliable_2020}, developing into a race between between attackers and defenders. Currently, the attack algorithm AutoAttack \cite{croce_reliable_2020} is used by the community to empirically evaluate adversarial robustness. While it does not provide a certification of robustness, it has been shown to bypass gradient masking in practice \cite{croce_reliable_2020}, and requires no hyper-parameter tuning. This makes it a strong public progress benchmark \cite{croce_robustbench_2020}. Following the current SOTA work in adversarial robustness \cite{liu_comprehensive_2023}, we use it in our work for all main evaluations.

\paragraph{Training robust models.}
\emph{Adversarial Training} emerged as the predominant paradigm to train robust models in practice \cite{madry_towards_2018, athalye_obfuscated_2018, dong_benchmarking_2020, salman_adversarially_2020, wong_fast_2020, liu_comprehensive_2023}. It is a complex bi-level algorithm, where at each iteration we generate a strong attack (via iterative optimization) and train the network to classify it correctly. In practice, such approaches, including the accelerated single-step Fast Gradient Sign Method (FGSM), require tuning many hyper-parameters and can be difficult to train \cite{wong_fast_2020,li_bag_2022, andriushchenko_understanding_2020,kim_understanding_2020}. Alternative approaches attempted to regularize models to have small input gradient norms \cite{seck_l_2019, liu_jacobian_2022, ross_improving_2017, jakubovitz_improving_2019, finlay_scaleable_2019, simon-gabriel_first-order_2019, lyu_unified_2015}. Despite strong theoretical arguments, no previous works have shown competitive performance on ImageNet from such regularizations. To our best knowledge, our work is the first to show its strong performance on modern architectures, and pinpoints activation function smoothness as the deciding factor of its effectiveness. 

\paragraph{Perceptually aligned gradients.} Previous works noted that robust models tend to have \textit{perceptually aligned gradients}, \ie, class gradients $\nabla_x f_\theta(x)_{y_t}$ that align with human perception  \cite{kaur_are_2019,santurkar_image_2019,srinivas_which_2023} (see also \Cref{fig:gradient-comparison}).
In this work, we consider the reverse implication and ask: do perceptually aligned gradients imply robustness? The work of \cite{ganz_perceptually_2023} aligned model class gradients $\nabla_xf_\theta(x)_{y_t}$ with several notions of class-representative images (defined via real images and generative models), and observed improved robustness on small datasets, but only small benefits on TinyImageNet. Our paper shows that simply aligning gradients to image edges yields much stronger robustness gains on the more challenging full ImageNet-1k. In addition to perceptual alignment, we also highlight several other properties that separate gradients of robust and non-robust models.

\section{Experimental Settings for Evaluating Robustness}
\label{sec:experimental-settings}


For all our experiments, we use the standard settings for large-scale experiments with modern architectures used by the community, following the work of Liu \etal \cite{liu_comprehensive_2023}, the current SOTA work in adversarial robustness, and the public RobustBench benchmark \cite{croce_robustbench_2020}. Specifically, we focus on a setting consisting of (1) the ImageNet-1K classification task, (2) Swin Vision Transformers architecture \cite{liu_swin_2021}, (3) the training recipe of Liu \etal \cite{liu_comprehensive_2023}, and (4) a full end-to-end knowledge adversary model with an $L_\infty$ constraint of $\frac{4}{255}$, empirically benchmarked with AutoAttack \cite{croce_reliable_2020}. See more details below.

\paragraph{Dataset}
Our work aims to provide insights into real-world, large-scale classification scenarios.  We choose to work with the challenging ImageNet-1K \cite{deng_imagenet_2009} dataset for all experiments. Additionally, the 224x224 resolution images of ImageNet enable higher-resolution visualization and easier analyses. 

\paragraph{Architecture}
We use the current state-of-the-art architecture for Adversarial Robustness on ImageNet, Swin Transformers \cite{liu_swin_2021}, albeit in base size. We will also start from a standard pre-trained checkpoint from \texttt{timm} \cite{wightman_pytorch_2019} in all experiments to reduce computational expense from 300 to 100 epochs per full training run (roughly 3 weeks to 1 week reduction on 8 V100 GPUs). From the work of Liu \etal \cite{liu_comprehensive_2023}, we know the impact of using pre-trained initializations and base size models is measurable but qualitatively equivalent, so we argue it is a sufficiently strong setting for our experiments. When analysing the effect of architectural choices in \cref{sec:effect-of-smoothness}, we run experiments on ResNet50's \cite{he_deep_2016} to allow for more extensive ablations, due to the aforementioned computational cost of training transformers.

\paragraph{Adversarial Training skyline and training recipe}
We use the work of Liu \etal \cite{liu_comprehensive_2023}, the current state-of-the-art for $L_\infty$ robust accuracy on ImageNet, both as a skyline to compare performance as well as for their strong training recipes. Unless otherwise stated, as per the work of Liu \etal \cite{liu_comprehensive_2023} all training recipes for Swin Transformers last 100 epochs, and use standard training tricks and augmentations like RandAugment \cite{cubuk_randaugment_2020, cubuk_autoaugment_2018}, mixup with label smoothing and random erasing \cite{zhang_mixup_2018}, the AdamW optimizer with weight decay \cite{loshchilov_decoupled_2017}, and model averaging \cite{izmailov_averaging_2018, gowal_uncovering_2020, bolme_average_2009}. Recipes used for ResNets are equivalent but halved in length. The full recipe details for all experiments are included in \cref{sec:training-details} of the appendix, and the code can be found on our \href{https://github.com/adriarm/robustness_input_gradients}{github}.

\paragraph{Attack benchmark}
We generally evaluate robustness using AutoAttack-$L_\infty$ \cite{croce_reliable_2020} with perturbation strength $\epsilon=\frac{4}{255}$, following the RobustBench standard \cite{croce_robustbench_2021}. AutoAttack has access to the end-to-end model, and contains a mix of white-box and black-box attacks. The white-box attacks themselves also use a mix of loss functions. When evaluating robustness for a dense set of perturbations strengths $\epsilon$, we use the PGD100 attack due to computational cost. Implementation-wise, we use the code and test set of Liu \etal \cite{liu_comprehensive_2023}. In addition to testing with AutoAttack, the work of Carlini \etal \cite{carlini_evaluating_2019} also provides a checklist of important practices, common pitfalls, and sanity checks in empirical evaluations of robustness, which we verify for Gradient Norm Regularization in the \cref{sec:robustness-checklist}.

\section{Small $L_1$ Gradient Norm Makes a Model Robust}


Mathematically, the stability of a function to small perturbations is captured by the norm of its derivative. Hence, it is a necessary condition for adversarial robustness to have small gradient norm \cite{andriushchenko_understanding_2020}. However, previous works encountered that minimizing gradient norm was not competitive with adversarial training on large benchmarks \cite{seck_l_2019, liu_jacobian_2022, ross_improving_2017, jakubovitz_improving_2019, finlay_scaleable_2019, simon-gabriel_first-order_2019, lyu_unified_2015}. Is this still the case with modern vision transformers \cite{liu_swin_2021}? Why did this performance gap exist? We take a deep look in this section.

A function stable to perturbations of the input has a small gradient \wrt the input. Empirically, it has been observed that adversarial robustness (obtained with adversarial training) \cite{madry_towards_2018} correlates with small gradient norm \cite{simon-gabriel_first-order_2019}. As we can see in \cref{tab:evidence-low-gradient-norms}, this continues to hold with state-of-the-art robust transformers. The expected $L_1$ norm of the loss-input gradient $\norm{\nabla_x \mathcal{L}} := \norm{\nabla_{\mathbf{x}}\mathcal{L}_{\text{CE}}(f_\theta(\mathbf{x},y)}_1$ is around two orders of magnitude smaller for robust models than for their non-robust counterparts of the same architecture. Furthermore, fixing the models and taking expectations over inputs conditioning on PGD10 attack success, the gradient is much smaller when the attack fails than when it succeeds (last two columns of \cref{tab:evidence-low-gradient-norms}).

\begin{table}[t]
    \caption{Accuracy, robustness, and gradient norm statistics on 10k ImageNet validation images for publicly available vulnerable and robust models from \texttt{timm} \cite{wightman_pytorch_2019} and RobustBench \cite{croce_robustbench_2021} respectively. The quantities Standard, AA, and PGD10 refer to clean accuracy, and AutoAttack and PGD10 robust accuracy respectively. The quantities $\mathbf{E}[L_1 | \protect\checkmark]$ and $\mathbf{E}[L_1 | \protect\crossmark]$ are the conditional expectations of the loss input-gradient $L_1$ norm conditioned on the PGD10 attack failing and succeeding respectively.}
    \label{tab:evidence-low-gradient-norms}
    \centering
    \resizebox{\linewidth}{!} {
    \begin{tabular}{cccccccc}
     &  & \multicolumn{3}{c}{\bf Accuracy} & \multicolumn{3}{c}{\bf Gradient Norm} \\ 
     \cmidrule(lr){3-5} \cmidrule(lr){6-8}
     {\bf Architecture} & {\bf Training} & Standard & AA & PGD10 & $\mathbf{E}[L_1]$ & $\mathbf{E}[L_1 | \checkmark]$ & $\mathbf{E}[L_1 | \crossmark]$  \\
    \midrule
    \multirow{2}{*}{Resnet-50} & Std. (He \etal) & 76.35 & 00.00 & 00.25 & 3013 & 0.030 & 3021 \\
     & Adv. (Salman \etal) & 63.99 & 34.96 & 39.98 & 56.82 & 11.78 & 86.82 \vspace{2mm}\\
    \multirow{2}{*}{Swin-B} & Std. (Liu \etal) & 84.84 & 00.00 & 02.77 & 3892 & 546 & 3983 \\
     & Adv. (Liu \etal) & 76.86 & 56.16 & 59.25 & 37.42 & 10.96 & 75.90 \vspace{2mm}\\     
    \multirow{2}{*}{Swin-L} & Std. (Liu \etal) & 86.13 & 00.00 & 02.07 & 2408 & 285.9 & 2453 \\
     & Adv. (Liu \etal) & 78.53 & 59.56 & 61.33 & 33.73 & 7.08  & 75.64 \\     
  \bottomrule
  \end{tabular}}
\end{table}

Despite the large amounts of theoretical and empirical evidence supporting low norms for robustness, the results of extensive previous work studying low gradient norms as a regularizer \cite{seck_l_2019, liu_jacobian_2022, ross_improving_2017, jakubovitz_improving_2019, finlay_scaleable_2019, simon-gabriel_first-order_2019, lyu_unified_2015} have so far been either inconclusive, or non-performant on large datasets like ImageNet. In the following section, we show how a simplified version of the gradient norm $L_1$ penalty, trained through double back-propagation \cite{drucker_improving_1992}, is within 5\% of the state-of-the-art on adversarial robustness on ImageNet despite seeing only natural examples, showing how small gradients have a larger driving role in robustness than previously thought. 

\subsection{Regularizing for Small Gradient Norms}
\label{sec:regularizing-for-small-gradient-norms}

Works in the literature studying gradient norm regularization have presented numerous formulations with differing details, such as optimizing Jacobian gradients \cite{jakubovitz_improving_2019, liu_jacobian_2022, seck_l_2019}, regularizing gradients of both natural and adversarial examples \cite{seck_l_2019}, or regularizing through a discrete scheme \cite{finlay_scaleable_2019}. In this work, we study the effect of the cleanest formulation possible of the objective, as presented in equation 4 of \cite{simon-gabriel_first-order_2019} and optimized in \cite{ross_improving_2017}, which we restate in \cref{eq:gradnorm-objective}

\begin{equation}
\label{eq:gradnorm-objective}
    \mathcal{L}_{\text{GradNorm}}(\mathbf{x},y) = \lambda_{\text{CE}}\mathcal{L}_{\text{CE}}(f_\theta(\mathbf{x}),y) + \lambda_{\text{GN}}\frac{\epsilon}{\sigma}\norm{\nabla_{\mathbf{x}}\mathcal{L}_{\text{CE}}(f_\theta(\mathbf{x}), y))}_1
\end{equation}
where $\mathcal{L}_{\text{CE}}$ is the cross-entropy loss, $\epsilon=\frac{4}{255}$ is the adversarial strength, $\sigma=0.225$ is the standard deviation used for normalization on ImageNet, and $\lambda_{\text{CE}}, \lambda_{\text{GN}}$ are weighing hyper-parameters set $0.8$ and $1.2$ respectively.


As we can see \cref{tab:grad-norm-autoattack-table}, training on the above objective yields a highly competitive model despite the constraints of seeing only natural examples and having 60\% of the computational budget. On AutoAttack $L_\infty$ with $\epsilon=\frac{4}{255}$, the standard benchmark for ImageNet \cite{croce_robustbench_2021}, gradient norm regularization obtains 51.58\% robust performance compared to the 56.12\% obtained by state-of-the-art adversarial training (also starting from a pretrained checkpoint) \cite{liu_comprehensive_2023}. For smaller epsilons, the gap shrinks to 1.18\% and 0.42\% for $\epsilon$ of 2 and 1 respectively, 
though it may be possible that the gap would be larger if the Adversarial Training was performed with a lower $\epsilon$ than $\frac{4}{255}$. 

\begin{table}[t]
    \caption{Robustness of a Swin Transformer trained with gradient norm regularization compared to natural training and state-of-the-art adversarial training on AutoAttack-$L_\infty$. Adversarial training performed from pretrained \texttt{timm} \cite{wightman_pytorch_2019} checkpoint using the recipe of \cite{liu_comprehensive_2023}.}
  \label{tab:grad-norm-autoattack-table}
  \centering
  \resizebox{.8\linewidth}{!}{
  \begin{tabular}{@{}lcccc@{}}
     & \multicolumn{1}{c}{\bf Clean} & \multicolumn{3}{c}{\bf AutoAttack-$L_\infty$}\\
     \cmidrule(lr){2-2} \cmidrule(lr){3-5} 
    {\bf Method} & - & $\epsilon=\frac{1}{255}$ & $\epsilon=\frac{2}{255}$ & $\epsilon=\frac{4}{255}$ \\    
    \midrule
    Natural Training  &  84.19 & 00.00 & 00.00 & 00.00 \\
    Grad. Norm ($\lambda_{\text{CE}}=0.8,\lambda_{\text{GN}}=1.2$) & 77.78 & 72.04 & 66.20 & 51.58 \\
    Adv. Train. (PGD-3, $\epsilon=4$) & 77.20 & 72.46 & 67.38 & 56.12\\
  \bottomrule
  \end{tabular}}
\end{table}

\begin{figure}[t!]
    \centering
    \includegraphics[width=\linewidth]{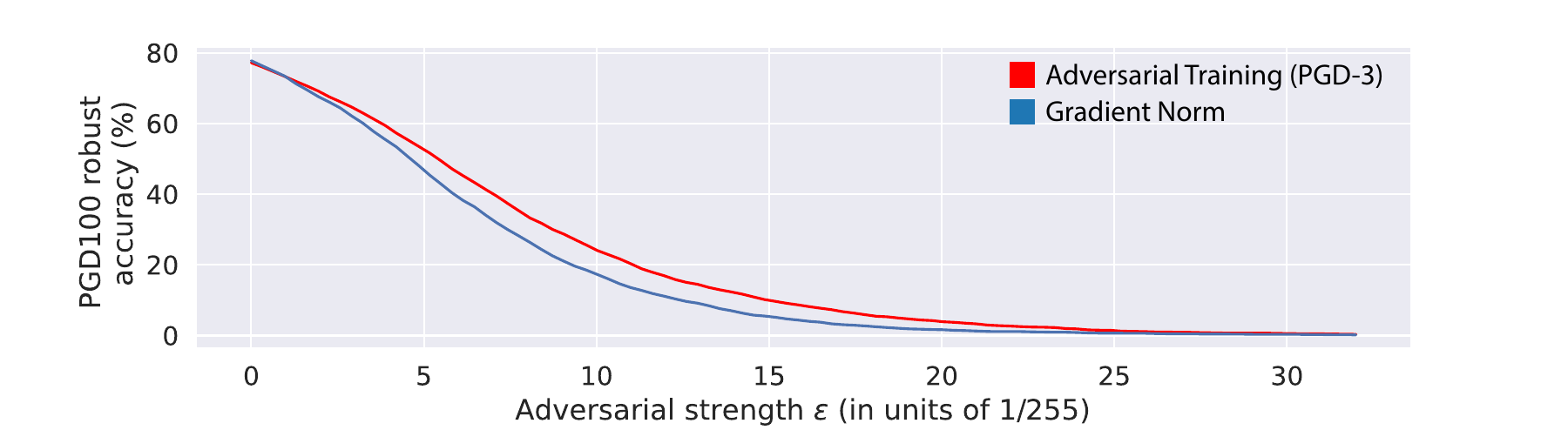}
    \caption{Robust accuracy vs epsilon for the PGD100 attack on ImageNet for Swin Transformer trained on Gradient Norm Regularization and state-of-the-art Adversarial Training. Gradient Norm Regularization achieves slightly better accuracy on clean images ($\epsilon = 0$) and good robust performance ($\epsilon > 0$), despite seeing only natural examples and having 60\% of the computational cost of Adversarial Training with PGD-3. Robust accuracy for both models smoothly converges towards 0\% as the adversarial strength grows.
    }
    \label{fig:grad-norm-pgd100-curve}
\end{figure}

More interestingly, we also evaluate behaviour for higher values of $\epsilon$. In \cref{fig:grad-norm-pgd100-curve}, we plot robust performance on PGD100 for a dense $\epsilon$ interval from 0 to $\frac{32}{255}$. 
As we can see, while the gap to adversarial training grows larger as a function of the adversarial strength $\epsilon$, it is always less than 10\%. Additionally, for $\epsilon<\frac{12}{255}$, 
accuracy is always two orders of magnitude above chance. This is an incredibly surprising result: despite gradient norm regularization working on only natural examples, it remains strong even for very large $\epsilon$. Moreover, while single-step adversarial training with FGSM will fail without injection of Gaussian noise prior to attack calculation \cite{wong_fast_2020} (a technique called random initialization), this is not necessary for gradient norm. Moreover, \cref{fig:grad-norm-pgd100-curve} also verifies the sanity check that robustness approaches 0\% as the adversarial strength $\epsilon$ grows unboundedly. 


This showcases both the strong bias towards smoothness endowed by modern architectural choices, as well as the strong robustness driving effect of small gradient norms. \cref{tab:grad-norm-computational-cost} displays computational cost comparisons between natural and adversarial training compared to gradient norm regularization. Roughly speaking, it means that for smooth swin transformers, adversarial training with PGD-3 is expending 40\% of its computational budget improving results by 5.1\% accuracy points, or about an 8.8\% relative increase. Similarly, minimizing the loss input-gradient $L_1$ norm is responsible for 92\% of state-of-the-art robust accuracy.

\begin{table}[t]
    \caption{Computational cost per batch comparison between natural training, adversarial training, and gradient norm regularization for Swin Transformer (base size). Theoretical cost measured in number of network passes per batch, and empirical cost measured in seconds per batch. Experiments conducted on the same set of $8$ V100 GPUs without mixed precision. Averages and standard deviations reported for the average batch time over three separate runs.}
  \label{tab:grad-norm-computational-cost}
  \centering
  \begin{tabular}{@{}lc@{\hskip 0.1cm}c@{\hskip 0.2cm}c@{\hskip 0.1cm}c}
    Method & \# passes & Rel. to Adv. Train. & Empirical cost ($s$) & Rel to Adv. Train. \\
    \midrule
    Nat. Train.  & 2 & 0.250 & 0.749 $\pm$ 0.00216 & 0.255 \\
    Grad. Norm. & 5 & 0.625 & 1.848 $\pm$ 0.01108 & 0.628 \\
    Adv. Train. (PGD-3) & 8 & 1.000 & 2.943 $\pm$ 0.00141 & 1.000 \\
  \bottomrule
  \end{tabular}
\end{table}

\begin{figure}[t!]
    \centering
    \includegraphics[width=\linewidth]{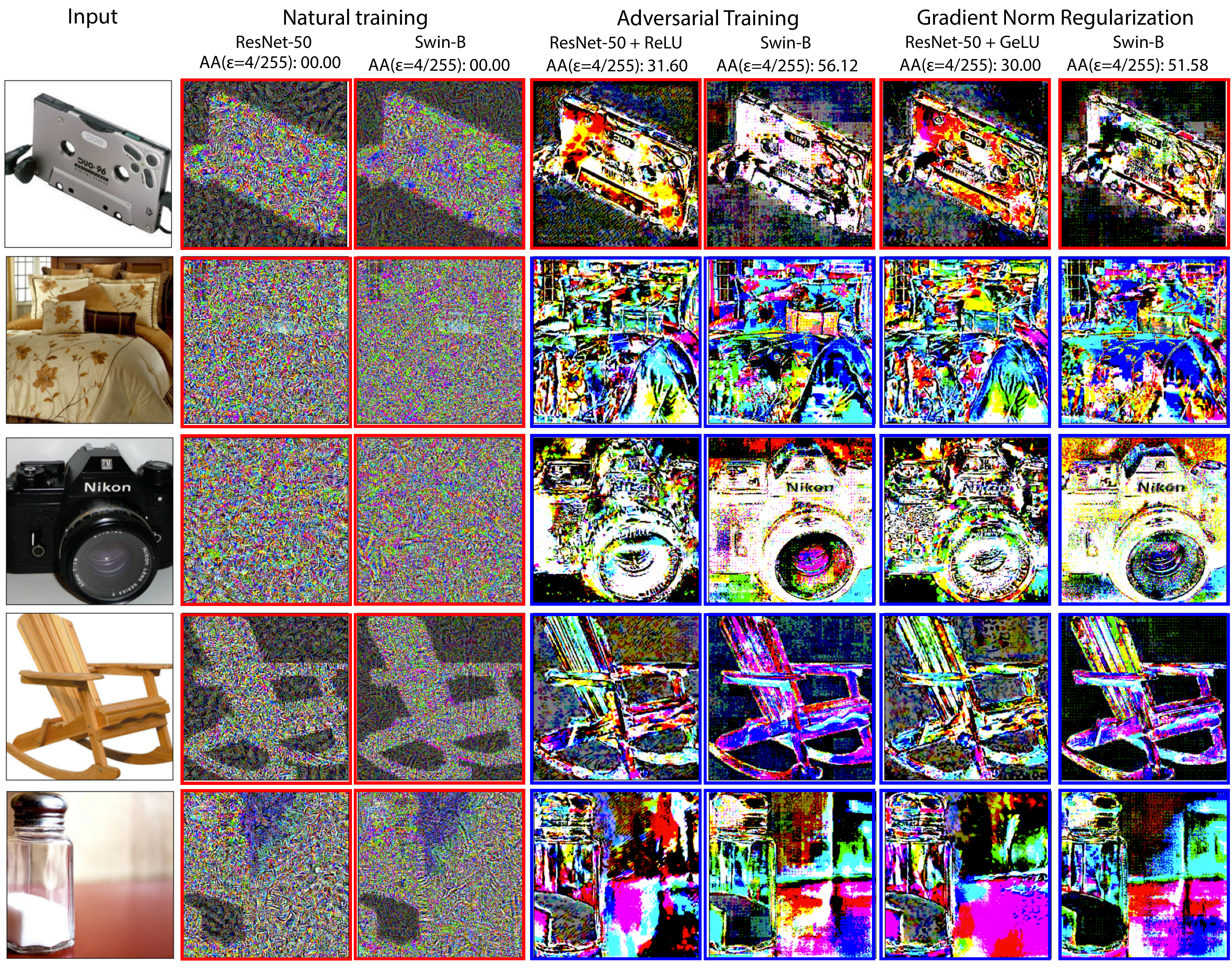}
    \caption{Comparison of PGD10 $L_\infty (\epsilon=\frac{4}{255})$ perturbations of non-robust and robust models across architectures for a set of images (same as in \cref{fig:gradient-comparison}). The border is blue if the model is robust to the perturbation, and red if it is not robust. As with clean input gradients, models can again be easily identified as vulnerable or robust simply by looking at the perturbations. Perturbations coming from robust models (adversarial training and gradient norm regularization) highly resemble the input images, though the visual similarity has decreased \wrt the input gradients. Perturbations originating from vulnerable models are now even more noise-like, with the exception of images with very flat backgrounds, potentially because the gradient may oscillate around zero in those areas. Perturbations normalized to [0, 1] for visualization purposes only. }
    \label{fig:attack-comparison}
\end{figure}

In the following section, we empirically show the drastic effect of smooth non-linearities on the performance of gradient norm regularization, even on smaller architectures such as ResNets.

\subsection{Smooth Activation Functions Make Gradient Norm Regularization Effective}
\label{sec:effect-of-smoothness}

The formulation of the gradient norm objective in \cref{eq:gradnorm-objective} is extremely similar to that of previous works \cite{seck_l_2019, finlay_scaleable_2019, ross_improving_2017, jakubovitz_improving_2019}, so why are results so different now? As was openly discussed by previous works \cite{finlay_scaleable_2019, simon-gabriel_first-order_2019}, ReLU networks are non-smooth, \ie, they have non-differentiable gradients. While they raised concerns regarding the effect of these non-differentiable gradients on gradient norm regularization, they did not conduct tests to measure the size of this effect, which we perform in this section.

\begin{figure}[t!]
    \centering
    \includegraphics[width=\linewidth]{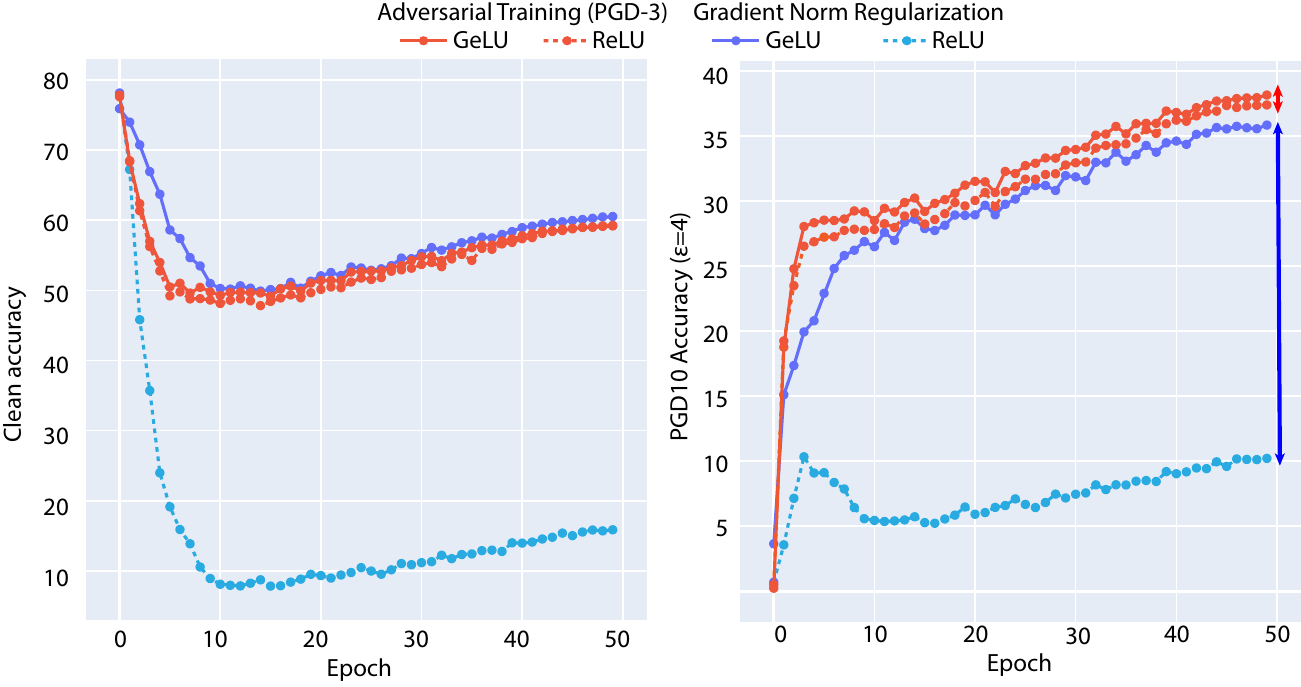}
    \caption{Clean and PGD10 ($\epsilon=4$) robust accuracy vs epoch for ResNet50 with ReLU and GeLU trained with Adversarial Training and Gradient Norm Regularization. We observe how the ReLU ResNet is not capable of handling the regularization objective at the appropriate strength. }
    \label{fig:resnet-performance-epoch} 
\end{figure}

We set up the following controlled comparison. First, we take a pre-trained ResNet-50  \cite{wightman_pytorch_2019} and replace all the ReLU non-linearities with smooth GeLUs and SiLUs \cite{hendrycks_gaussian_2023}. This change causes clean accuracy to decrease to 0\%, so we finetune over three epochs. For consistency, we also finetune the ResNet-50 with ReLUs with the same recipe and seed. Next, we make a copy of each network and train each copy on the same recipe (a halved version of the recipe used for transformers), with both adversarial training with PGD-3 and gradient norm regularization. Throughout all the trainings we keep the batch-normalizations in evaluation mode in order to isolate the effect of the non-linearity. The evaluations on $L_\infty$-AutoAttack($\epsilon=\frac{4}{255}$) are shown on \cref{tab:resnet50-autoattack}. We show validation accuracies as functions of epoch in \cref{fig:resnet-performance-epoch}.

\begin{table}[t]
    \caption{Clean and $L_\infty$-AutoAttack accuracy for ResNet50 with ReLU, GeLU, and SiLU non-linearities trained with both Adversarial Training and GradNorm for 50 epochs using a shortened version of the Adversarial Transformer recipe of \cite{liu_comprehensive_2023}.}
  \label{tab:resnet50-autoattack}
  \centering
  \begin{tabular}{cc|cc}
    \multicolumn{2}{c|}{\bf Method} & \multicolumn{2}{c}{\bf Accuracy} \\
     \cmidrule(lr){1-2}\cmidrule(lr){3-4}
     Training & Non-linearity & Standard & AutoAttack-$L_\infty$\\
    \midrule
    \multirow{3}{*}{GradNorm} & ReLU & 16.94 & 6.82 \\
     & GeLU & 60.34 & 30.00 \\
     & SiLU & 61.84 & 30.58 \\
    \cdashlinelr{1-4}
    \multirow{3}{*}{Adv. Train.} & ReLU & 59.46 & 31.60 \\
    & GeLU & 59.34 & 32.64 \\
    & SiLU & 60.58 & 33.40 \\
  \bottomrule
  \end{tabular}
\end{table}

As we can see in \cref{tab:resnet50-autoattack} and \cref{fig:resnet-performance-epoch}, the ResNet with ReLU is completely incapable of properly fitting the objective at the appropriate strength, with clean performance sharply decaying and robust performance barely increasing, compared to the ResNet with GeLU, despite training on the same recipe with the same regularization objective and weights. In contrast, the gradient norm regularized GeLU ResNet displays similar convergence behaviour to the adversarially trained model, obtaining extremely similar final clean and robust accuracies.

The work of \cite{xie_smooth_2021} conducted a similar analysis for Adversarial Training, observing small increases in performance from using smooth non-linearities. As we can see from \cref{tab:resnet50-autoattack}, for Gradient Norm regularization the effect is more than 20 times larger: robust performance on AutoAttack with adversarial training increases by 1\%, while for gradient norm the increase is roughly 23\%. In essence, the gradient norm regularization objective minimizes a penalty on the gradients of the network; since for ReLU networks the latter is non-differentiable, Taylor's theorem does not necessarily hold on the gradient loss, so there is no guarantee that gradient descent will work.

Additionally, we found the usage of adaptive optimizers like Adam \cite{kingma_adam_2017}, as well as a relatively small warm-up learning rate of $10^{-5}$, to be extremely important. Especially at the beginning, the size of the norm of the gradient changes drastically; using a non-adaptive optimizer or too high a learning rate also causes performance to similarly crash, even on smooth networks, for the regularization weights required to reach performance comparable to adversarial training. This may have been another reason behind the lack of conclusive success of previous works.

In \cref{fig:attack-comparison}, we visualize PGD10 perturbations of the same vulnerable and robust models from \cref{fig:gradient-comparison}. Similarly as in \cref{fig:gradient-comparison}, obvious visual differences exist between the perturbations of vulnerable and robust models.

\begin{figure}[t]
    \centering
    \includegraphics[width=\linewidth]{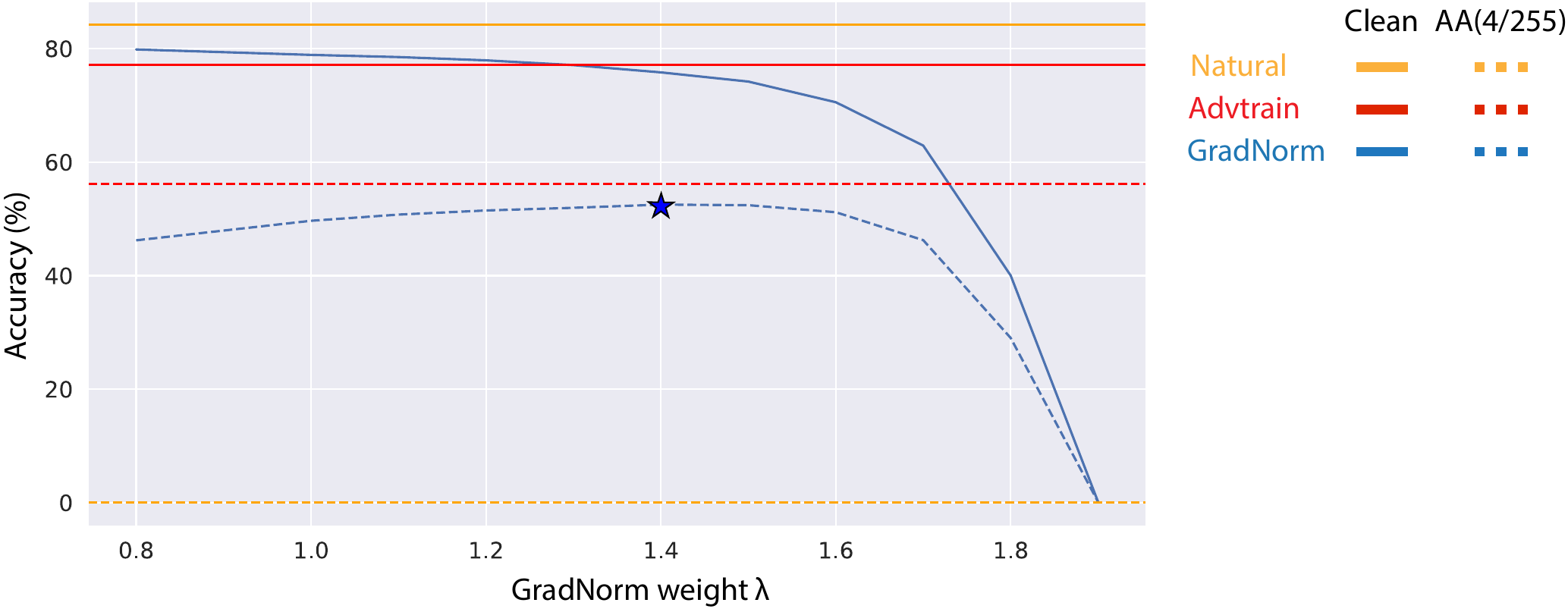}
   \caption{Clean and AutoAttack (AA; $\epsilon=\frac{4}{255}$) accuracy for optimizing $(2 - \lambda) \mathcal{L}_\text{CE} + \lambda \lVert\nabla_x \mathcal{L}\rVert_1$ for Swin Transformer. Models obtained by finetuning original GradNorm model from \cref{tab:grad-norm-autoattack-table} for 30 epochs due to computational cost of training from scratch. Maximum robust accuracy obtained for $\lambda=1.4$, which has clean accuracy 76.28 and AutoAttack accuracy 52.48. The robust accuracy gap to Adversarial Training is 3.64\%.}
    \label{fig:pareto}
\end{figure}
\subsection{Trading Off Smoothness and Performance}

\cref{fig:pareto} plots clean and AutoAttack \cite{croce_reliable_2020} accuracy as a function of GradNorm weight $\lambda$ for Swin Transformer. Increasing the GradNorm weight from 1.2 to 1.4 increases AutoAttack accuracy at $\epsilon=\frac{4}{255}$ by almost 1\%, from 51.58\% to 52.48\%. While the gap to Adversarial Training robust accuracy is reduced to 3.64\%, clean accuracy degrades by 1.5\%, from 77.78\% to 76.20\%. Hence, the "optimal" value depends on the user-specified trade-off between clean accuracy and robustness.

\begin{figure}[t]
    \centering
    \includegraphics[width=\linewidth]{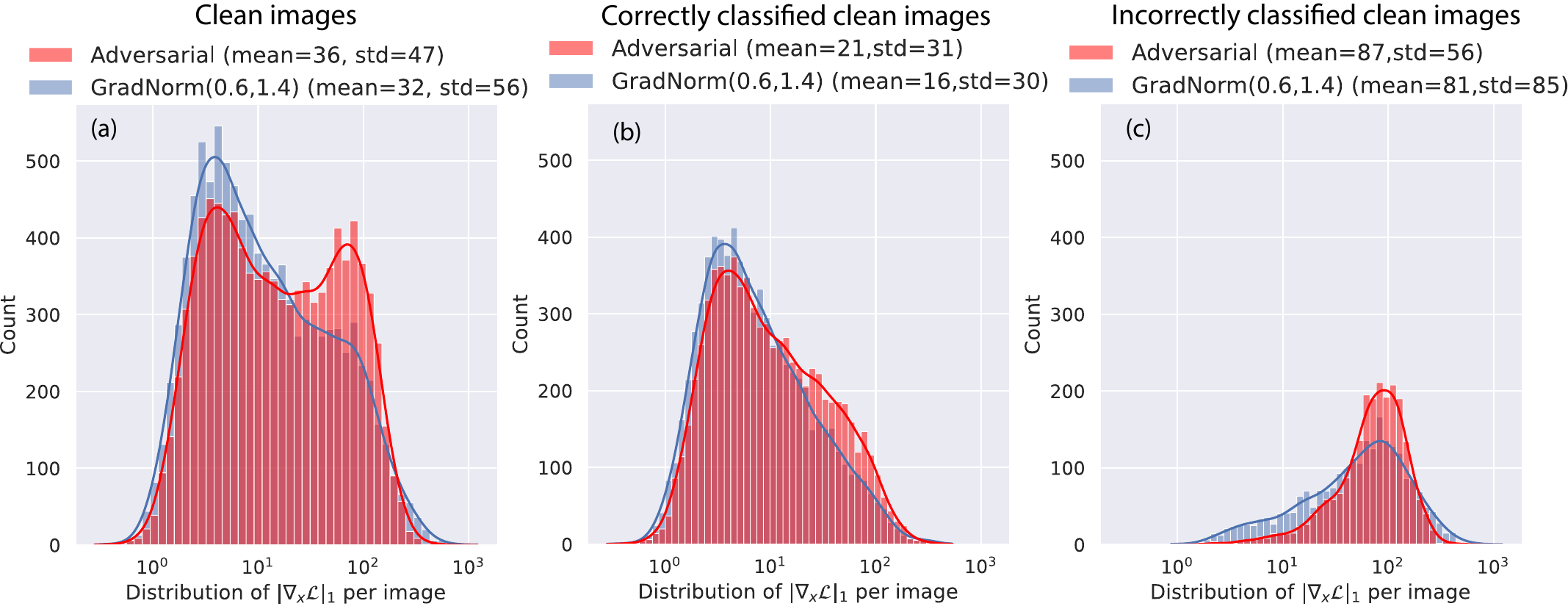}
    \caption{$L_1$ loss-input gradient norm ($\norm{\nabla_x \mathcal{L})}_1$) distribution of AdvTrain and best GradNorm Swin Transformer models for different sets of clearn images. 
    On all images (\textbf{left}), the adversarially trained model has a bimodal distribution, while the GradNorm model does not. This is mostly due to differing behaviour on incorrectly classified examples (\textbf{right}). While the AdvTrain model has a highly concentrated distribution on large gradient norms for incorrect examples, the GradNorm model has a much higher proportion of low gradient norms.} 
    \label{fig:gradient-norm-distribution}
\end{figure}
\begin{figure}[t]
    \centering
    \includegraphics[width=\linewidth]{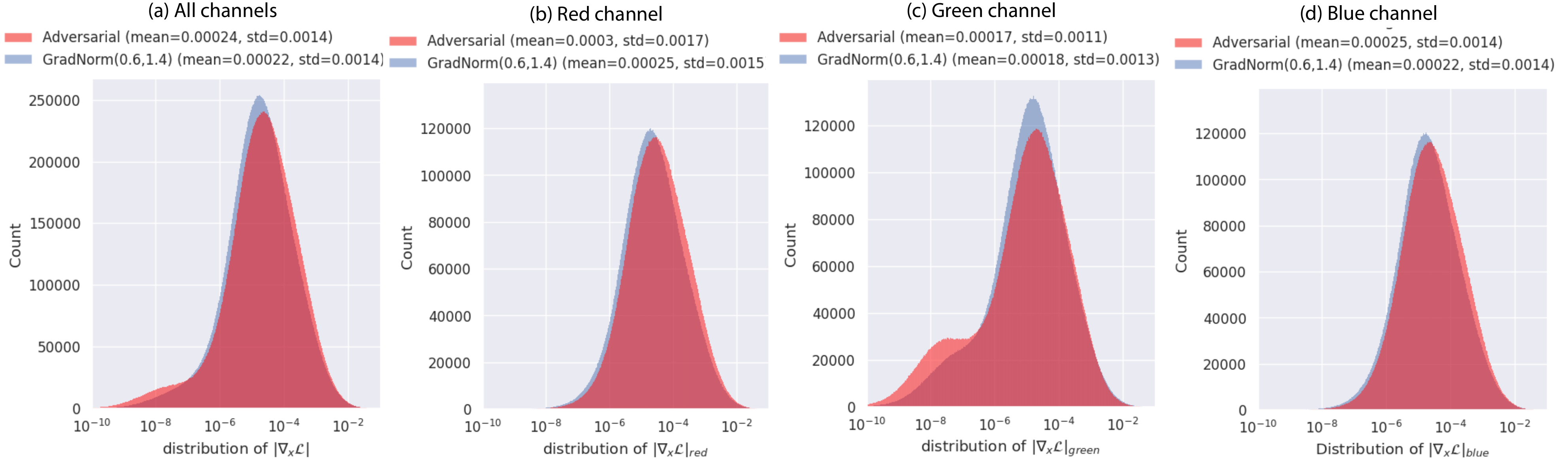}
    \caption{Distribution of absolute value of gradients over 1k images over all channels  (\textbf{a}), and each RGB color channel (\textbf{b,c,d}), for AdvTrain and best GradNorm Swin Transformer models. In \textbf{a}, we see adversarial gradients have a significantly fatter left tail; essentially, the small values are much lower. From the \textbf{b,c,d} channel-specifcic plots, we observe that most of this difference is owed to the green channel: the small values of the green channel of adversarially trained gradients are very small. This asymmetric behaviour is partially missing from the gradient norm regularized models.
    }
    \label{fig:log-absolute-value-gradients}
\end{figure}

\section{Beyond Small $L_1$ Norm: Other Properties of Robust Gradients }
\label{sec:beyond-small-l1-norm}


In this section, we investigate properties of adversarially robust gradients beyond small average $L_1$ gradient norm: (1) differences between gradients of correctly and incorrectly classified clean examples, (2) differences between RGB channels of gradients, and (3) alignment of gradients to image edges.


\subsection{Image Gradient Norm Correlates with Clean Classification Accuracy}

\cref{fig:gradient-norm-distribution} plots the $L_1$ norm distribution of loss-input gradients for clean images for the adversarially trained model, and the most robust GradNorm model of \cref{fig:pareto} ($\lambda=1.4$). The adversarially trained model exhibits strong bimodality (\cref{fig:gradient-norm-distribution}.a). This is due to differing behaviour on wrongly-classified clean examples, as we can see in \cref{fig:gradient-norm-distribution}.c. The gradient norm distribution for incorrectly classified clean examples is highly concentrated around large values for the adversarially trained model, but is more distributed around low values in the GradNorm model. 

\subsection{Not All Color Channels Have Equally Small Gradient Norms}

\cref{fig:log-absolute-value-gradients} plots the distribution of the absolute values of gradients across 1000 validation images. In \cref{fig:log-absolute-value-gradients} left, we see that adversarially trained gradients have more small values than gradient norm regularized gradients, even though the mean is slightly higher. The distribution for the latter tapers off at $10^{-9}$, while the former has a bump between $10^{-9}$ and $10^{-10}$. In other words, the adversarially trained gradients are sparser, with more dimensions closer to zero. But what are these additional sparse dimensions?

The right three plots of \cref{fig:log-absolute-value-gradients} show the gradient magnitude distributions \emph{for each channel}. 
we observe that the difference observed in above is almost entirely due to the green channel of the gradients. This asymmetric role of the green channel is seemingly unique to adversarial training. It's possible this may be due to the special role that the green channel plays in image coding, such as sampling using Bayer filters \cite{lukac_color_2005}, but we are not able to test it experimentally without access to ImageNet in RAW format. Instead, we conducted alternative experiments where GradNorm is penalized stronger for the green channel, but didn’t obtain positive results. Doubling the weight of the green channel reduced both clean and robust accuracy, from 77.74\% to 77.38\% and 51.66\% to 51.38\% respectively.

\subsection{Aligning Gradient to Image Edges Improves Robustness}
\label{sec:edge-regularization}

\begin{figure}[t]
    \centering
    \includegraphics[width=\linewidth,height=\textheight,keepaspectratio]{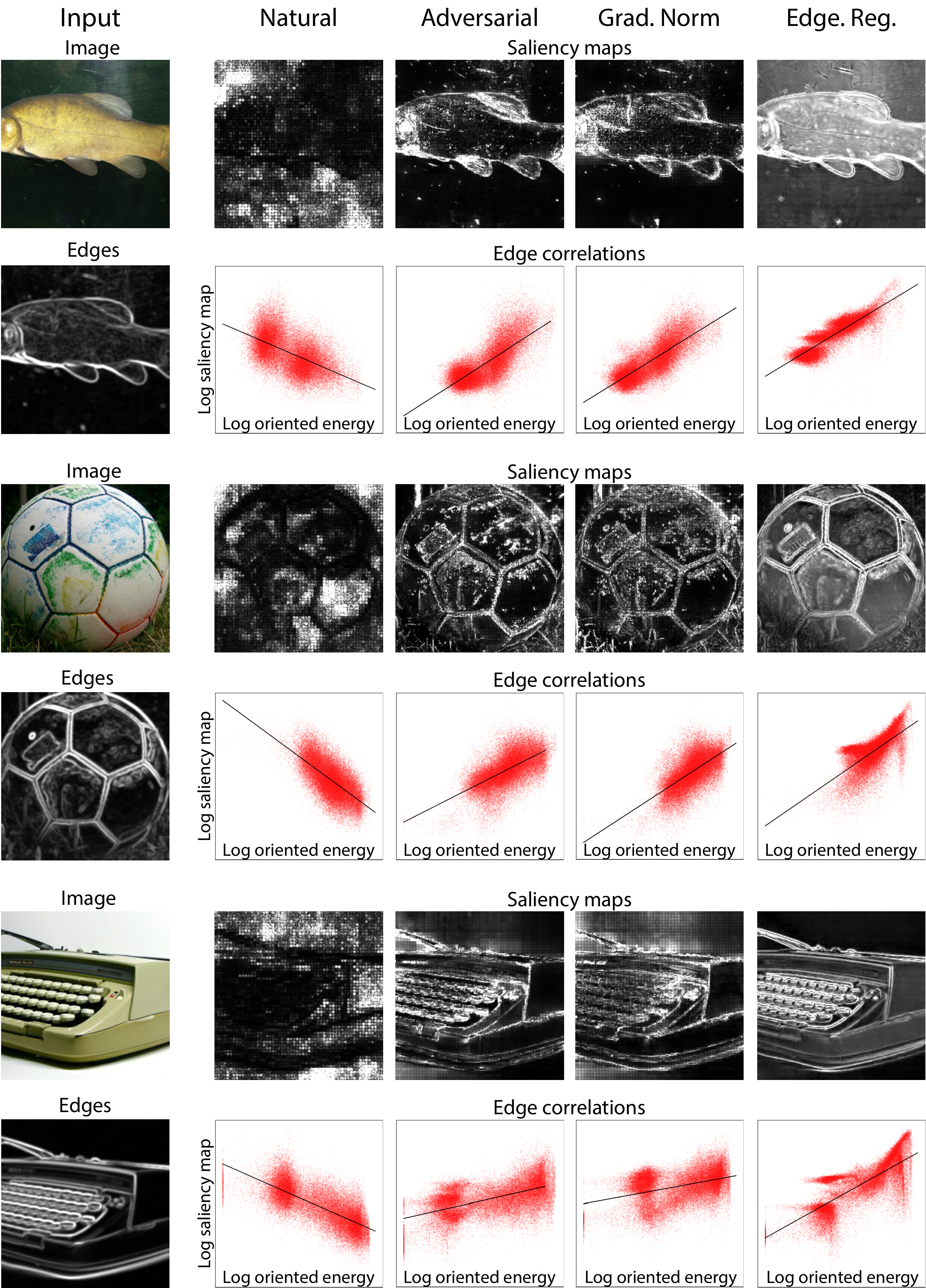}
    \caption{Scatter plots of log gradient magnitude vs log oriented energy for a non-robust and robust Swin Transformer. Oriented energy is calculated as $\text{edge}(x) = |g_u * x|^2 + |g_v * x|^2$, where $*$ denotes the convolution operation, and $g_u,g_v$ denote horizontal and vertical Sobel derivative filters respectively. Oriented energy is clamped at 1e-3 to eliminate extremely low outliers. Saliency maps are clipped to percentile 0.95 for visualization purposes only.}
    \label{fig:gradient-edges}
\end{figure}

Adversarially robust models are known to be perceptually aligned---class gradients $\nabla_x f_\theta(x)_{y_t}$ are image-like and align with human perception  \cite{kaur_are_2019,santurkar_image_2019,srinivas_which_2023}. In particular, the saliency maps (the per-pixel maximum absolute value of the target class input-gradients) of vulnerable and robust models display opposite correlations with the image edges despite highlighting the same object. This is seen visually in \cref{fig:gradient-edges} and numerically in \cref{tab:class-gradient-correlation}. The image edges, also known as the oriented energy of the image, measure of the amount of local change at each pixel. We calculate them using Sobel filters \cite{freeman_design_1991} as is standard in signal processing. In simple terms, horizontal and vertical derivatives are calculated using a convolution, squared, and added.

\cref{tab:class-gradient-correlation} reports the log-log correlation between both the saliency map and the loss input-gradient absolute value with the edges of the input. For the naturally trained model, both values are significantly negative at around -0.45, for both robust models, they are significantly positive at around +0.56.

In \cref{fig:gradient-edges}, this similarity between saliency maps and image edges (\ie, oriented energy) is visualized, showing the correlation between log saliency map and log oriented energy for individual images. Like in \cref{tab:class-gradient-correlation}, we observe strong positive correlation for robust models, and weaker negative correlations for non-robust models.

\begin{table}[]
\caption{Average and standard deviation of Pearson correlation between log saliency map magnitude and log oriented energy across 10k validation images. We remark a positive correlation of $+0.56$ for the robust models and a negative correlation of $-0.45$ for vulnerable vanilla models. This shows that their gradients are spatially located in opposite regions of the image, the edge regions vs the flat regions respectively, despite accurately classifying the same object. Moreover, we believe these values undersell the relationship, as the edges are naively calculated over the whole image and thus highlight irrelevant objects. Edges / oriented energy calculated as $\text{edge}(x) = |g_u * x|^2 + |g_v * x|^2$, where $*$ denotes the convolution operation, and $g_u,g_v$ denote Sobel horizontal and vertical derivative filters respectively.}
  \label{tab:class-gradient-correlation}
  \centering
  \resizebox{\linewidth}{!}{
  \begin{tabular}{l@{\hskip 0.8cm}c@{\hskip 0.1cm}ccc}
    & \multicolumn{2}{c}{Accuracy} & \multicolumn{2}{c}
    {Pearson corr. of log w/ log oriented energy} \\
    \cmidrule(lr){2-3} \cmidrule(lr){4-5}
    Training & Clean & AutoAttack & Saliency map & $\left|\nabla_x\mathcal{L}_{\text{CE}}(f_{\theta}(x), y)\right|$ \\
    \midrule
    Natural & $\mathbf{84.19}$ & $00.00$ & $-0.4510 \pm (0.1467)$ & $-0.4428 \pm (0.1498)$ \\
    Gradient Norm & $77.78$ & $51.58$ & $+0.5627 \pm (0.1640)$ & $+0.5679 \pm (0.1647)$ \\
    Adversarial (PGD-3) & $77.20$ & $\mathbf{56.16}$ & $+0.5627 \pm (0.1569)$ & $\mathbf{+0.5692 \pm (0.1569)}$ \\
    \cdashlinelr{1-5}
    Edge regularization & $76.80$ & $35.02$ & $\mathbf{+0.6055 \pm (0.1644)}$ & $+0.3882 \pm (0.2020)$ \\
  \bottomrule
  \end{tabular}}
\end{table}

A natural question that arises is whether this property is merely a consequence of robustness or actually induces robust behavior. We find that, to a significant extent, it is actually the latter. Aligning class gradients ($\nabla_x f_\theta(x)_{y_t}$, \ie, the gradient of target-class logit \wrt input) with image edges yields 60\% of the robustness of SOTA Adversarial Training. This is reported in \cref{tab:class-gradient-correlation} under the name ``Edge Regularization''. The exact form of the regularizer is given by the following \cref{eq:sobel-objective}.
\begin{alignat}{3}
      \label{eq:sobel-objective}
    \mathcal{L}_{\text{edge}}(\mathbf{x},y) \coloneqq \mathcal{L}_{\text{cos}}(\nabla_{\mathbf{x}}f_\theta(x)_t, |g_u * x|^2 + |g_v * x|^2) \\[0.7ex]
    t \sim \text{softmax}\left(\frac{f_\theta(x)}{0.5}\right)  
\end{alignat}
where $\mathcal{L}_{\text{cos}}$ is the cosine similarity loss and $\nabla_{\mathbf{x}}f_\theta(x)_y$ is a class gradient. The class is chosen according to the probabilities predicted by the model modulated by a temperature of 0.5, expressed mathematically with $\text{softmax}(f_\theta(x)/0.5)$. This is done to bias selections towards the likely classes. Additionally, $|g_u * x|^2 + |g_v * x|^2$ are the image edges: $*$ denotes 2D convolution, and $g_u,g_v$ denote gaussian horizontal and vertical derivative filters respectively.

In addition to optimizing the class gradient, we also attempted other forms of edge regularization, such as using loss gradients or the saliency map, but were not able to obtain good results. Please see \cref{sec:edge-regularization-appendix} for results with alternate losses and other training details.


Unlike gradient norm regularization, \cref{eq:sobel-objective} does nothing to a-priori regularize the norm of the gradients, only the direction. While not at the level of models regularized purposefully for robustness, this serves as significant evidence to support the claim that perceptual similarity directly causes robustness. Moreover, the true value of the result may lie in the following: it is much easier to conceptualize devising an architecture that natively focuses on edges than one that has low gradient norm or is resistant to arbitrary perturbations. That is, the success of edge regularization could potentially provide a start towards a \emph{structurally robust} architecture \ie one that displays adversarial robustness even when trained normally. We leave this direction to future work.

\section{Implications}

What properties of input gradients characterize model robustness? In our work, we find that, on architectures with smooth non-linearities, cleanly minimizing the $L_1$ norm of the loss input gradients achieves close to state-of-the-art robust performance, despite never training on perturbed examples. This implies that (1) model robustness is significantly characterizable by behaviour on natural inputs, and (2) architectural changes like non-linearity choice may drastically change the effectiveness of alternate approaches towards robustness. We also find an additional characterization of robust gradients based on image edges, independent of norm, that achieves $60\%$ performance of state-of-the-art. While numerically weaker than the gradient norm result, the implications are still potentially significant. Specifically, it provides a possible hint towards a \textit{naturally robust architecture} that, through directly enforcing dependence only on the edge regions of the image, is resistant to perturbations despite being naturally trained.

Our results show that input gradients provide a useful lens to both analyze and improve model robustness. This approach suggests several interesting future research directions to explore: (1) Are there other model properties that quantify robustness? (2) How can we characterize the performance gap between Gradient Norm Regularization and Adversarial Training? (3) How do these properties explain the relationship between robustness and architecture in general (beyond activation smoothness)? Based on our results, we believe that answering these questions will lead to further insights and advances in building more general and robust deep models.



\paragraph{Broader Impact} Deep neural networks have become the gold standard for computer vision tasks, but are also extremely brittle under small distribution shifts, including adversarial attacks. In this work, we highlighted deep connections between the robustness of neural networks and the statistics of their gradients, and proposed several approaches to train robust and highly-performant neural networks. We believe that our findings are important for applying such models in real-world safety-critical tasks.

\newpage
\section*{Acknowledgements}
Adrián Rodríguez-Muñoz is supported by the LaCaixa fellowship. Tongzhou Wang is supported by the ONR MURI program.

\bibliographystyle{splncs04}
\bibliography{bib}

\newpage
\appendix

\section[A]{Second-Order Analysis}
\label{sec:analysis}

Our main paper focuses on analyses of first-order statistics involving input-gradients. In this section, we provide additional analyses focusing on second-order behaviors, following previous work \cite{rocamora_efficient_2023, srinivas_efficient_nodate, moosavi-dezfooli_robustness_2018, tsiligkaridis_second_2020, qin_adversarial_2019, singla_low_2021}. In particular:
\begin{itemize}
    \item \cref{sec:geometry} measures the geometry statistics introduced by Srinivas \etal \cite{srinivas_efficient_nodate}.
    \item \cref{sec:linerr} measures the local linearity error of Rocamora \etal \cite{rocamora_efficient_2023}.
    \item  \cref{sec:attackdir} measures loss $\mathcal{L}$, $L_1$ gradient norm $|\nabla\mathcal{L}|_1$, and normalized curvature $\frac{|\nabla^2\mathcal{L}|_2}{|\nabla\mathcal{L}|_2}$ \cite{srinivas_efficient_nodate} as we move in the attack direction.
\end{itemize}
In contrast to previous works on smaller datasets, our analyses focus on large-scale models trained on ImageNet.

\subsection{Geometry Statistics}
\label{sec:geometry}

\cref{tab:curvature} plots average and standard deviation of the geometry statistics introduced by Srinivas \etal \cite{srinivas_efficient_nodate}: the loss-input gradient $L_2$ norm $|\nabla\mathcal{L}|_2$, the loss-input hessian spectral norm $|\nabla^2\mathcal{L}|_2$, and the normalized curvature $C_\mathcal{L} \coloneqq \frac{|\nabla^2\mathcal{L}|_2}{|\nabla\mathcal{L}|_2}$.

\begin{figure}[t]
    \centering
    \includegraphics[width=\linewidth]{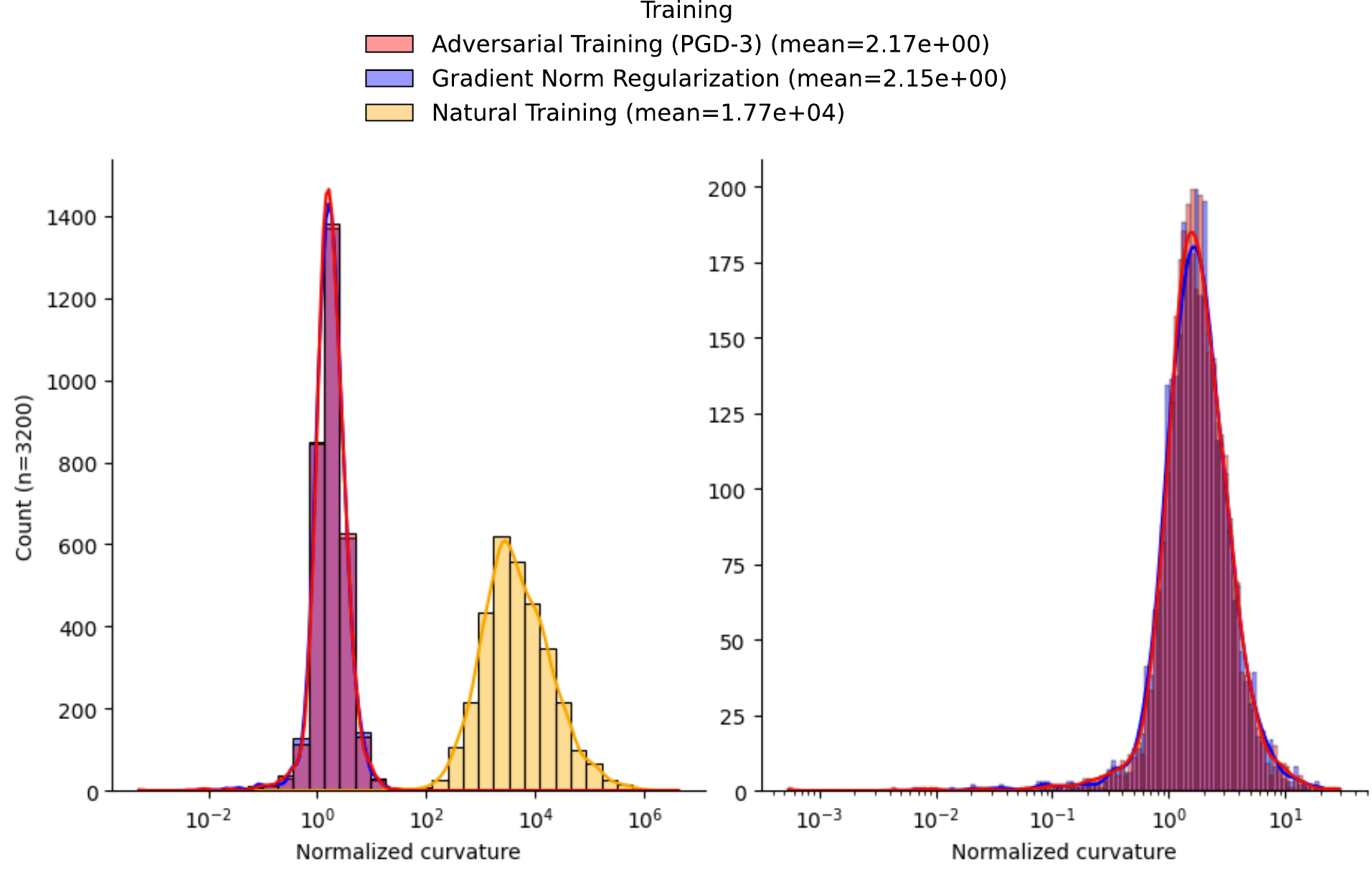}
    \caption{Distribution of normalized curvature \cite{srinivas_efficient_nodate} over 3200 ImageNet validation images, (a) plots all three models and (b) plots just the robust models. Due to the high variation in scale over images, the graph is plotted with the x-axis in log scale. In (a) we see robust training (both gradient norm regularization and adversarial training) leads to an improvement in normalized curvature of over three orders of magnitude \wrt to natural training. Surprisingly, amongst the robust models adversarial training has the slightly higher average curvature, though the distributions are so close they cannot be distinguished in the graph. Architecture is Swin Transformer. 
    }
    \label{fig:curvature}
\end{figure}

Standard deviations are very high due to a high variability in scale across examples, which we think is a feature unique to ImageNet as this was not observed by Srinivas \etal, though it could also be due to the much larger transformer model. Hence, we plot the distribution of curvature across examples in log scale in \cref{fig:curvature}.

As we can see, robust training leads to a decrease in normalized curvature of over three orders of magnitude \wrt natural training; hessian spectral norms drop by more than six, while gradient $L_2$ norms by around two. Between the two robust models, all geometry statistic averages are surprisingly very similar, with the normalized curvature of the adversarially trained model actually being slightly higher. However, we also observe that the standard deviation of gradient and hessian norm are significantly higher for gradient norm regularization, despite normalized curvature standard deviation being lower; this means that gradient and hessian norms vary on the same examples, such that normalized curvature remains small. Through visual examination of \cref{fig:curvature}.b, we see that the curvature distribution is also very similar across the two robust models. A possible explanation is that the gradient norm regularized model overfits to the clean examples, and thus displays similar numbers to the adversarially trained model despite its lower performance.


\begin{table}[t]
\caption{Geometry statistics as presented in Srinivas \etal \cite{srinivas_efficient_nodate}. Due to the high variation of the statistic scale across images standard deviations are very high, especially for the natural model. Hence, we also show the distribution of the curvature in \cref{fig:curvature}. Robust training (both gradient norm regularization and adversarial training) leads to an improvement in normalized curvature of over three orders of magnitude \wrt to natural training. Surprisingly, amongst the two robust models adversarial training has the highest curvature, though the numbers are quite close. Statistics calculated over 3200 ImageNet validation images.}
  \label{tab:curvature}
  \centering
  \resizebox{\linewidth}{!}{
  \begin{tabular}{l@{\hskip 0.1cm}c@{\hskip 0.1cm}cccc}
    & \multicolumn{2}{c}{Accuracy} & \multicolumn{3}{c}
    {Geometry} \\
    \cmidrule(lr){2-3} \cmidrule(lr){4-6}
    Training & Clean & AutoAttack & $\mathbb{E}_{\mathbf{x}}\norm{\nabla \mathcal{L}(\mathbf{x})}_2$ & $\mathbb{E}_{\mathbf{x}}\norm{\nabla^2 \mathcal{L}(\mathbf{x})}_2$ & $\mathbb{E}_{\mathbf{x}}C_\mathcal{L}(\mathbf{x})$ \\
    \midrule
    Natural training & $\mathbf{84.19}$ & $00.00$ & $28.7 \pm (93.8)$  & $10^6 \pm (10^7)$ & $10^4 \pm (10^5)$  \\
    Gradient norm regularization & $77.78$ & $51.58$ & $0.31 \pm (0.67)$ & $0.79 \pm 3.25$  & $2.15 \pm (1.72)$  \\
    Adversarial training (PGD-3) & $77.20$ & $\mathbf{56.16}$ & $0.31 \pm (0.44)$ & $0.75 \pm 1.71$ & $2.17 \pm (1.83)$ \\
  \bottomrule
  \end{tabular}
  }
\end{table}

\subsection{Local Linearity Error}
\label{sec:linerr}

As defined by Rocamora \etal in \cite{rocamora_efficient_2023}, minimizing the local linearity error optimizes the linearity of the model \wrt uniformly distributed perturbations. We rewrite their objective in \cref{eq:linerr}
\begin{alignat}{3}
\label{eq:linerr}    \mathbf{E}_{\alpha,\eta_1,\eta_2}&\left[|\alpha\mathcal{L}(x+\eta_1) + (1-\alpha)\mathcal{L}(x+\eta_2) - \mathcal{L}(x+\alpha\eta_1+(1-\alpha)\eta_2)|^2\right] \\ &\alpha\sim U(0,1) \hspace{0.2cm} \eta_1,\eta_2\sim U(-\epsilon,\epsilon)
\end{alignat}
where $\epsilon=4./255.$

\cref{tab:linerr} reports mean and standard deviation of the local linearity error across 10000 ImageNet validation images. As we can see, the two robust models have significantly smaller local linearity errors by about two orders of magnitude compared to the naturally trained model. Within the two robust models, adversarial training has lower average local linearity error, consistent with its superior robustness.

\begin{table}[t]
\caption{Average and standard deviation of the local linearity error of Rocamora \etal \cite{rocamora_efficient_2023}. Due to the high variation of the scale across images, standard deviations are larger than the mean. Hence, we also report mean and standard deviation of the base 10 logarithm of the error. We add $10^{-17}$ before computing log statistics for numerical stability. Robust training (both gradient norm regularization and adversarial training) leads to an improvement in local linearity error of two orders of magnitude \wrt to natural training. Within the robust models, adversarial training has significantly lower local linearity error, reflecting its superior robustness and second-order stability. Statistics calculated for 10000 ImageNet validation images. Architecture is Swin Transformer.}
  \label{tab:linerr}
  \centering
\resizebox{\linewidth}{!}{
  \begin{tabular}{l@{\hskip 0.1cm}c@{\hskip 0.1cm}cc@{\hskip 0.2cm}c}
    & \multicolumn{2}{c}{Accuracy} & \multicolumn{2}{c}
    {Local linearity error} \\
    \cmidrule(lr){2-3} \cmidrule(lr){4-5}
    Training & Clean & AutoAttack & linerr & $\log_{10}(\text{linerr})$ \\
    \midrule
    Natural Training & $\mathbf{84.19}$ & $00.00$ & $\num{3.55e-3} \pm (\num{2.69e-2})$ & $-4.57 \pm (1.63)$ \\
    Gradient norm regularization & $77.78$ & $51.58$ & $\num{1.86e-5} \pm (\num{1.65e-4})$ & $-6.60 \pm (1.43)$ \\
    Adversarial training (PGD-3) & $77.20$ & $\mathbf{56.16}$ & $\mathbf{\num{8.2e-6} \pm (\num{5.08e-5})}$ & $\mathbf{-6.84 \pm (1.48)}$ \\
  \bottomrule
  \end{tabular}
  }
\end{table}

\begin{figure}[t]
    \centering
    \includegraphics[width=\linewidth]{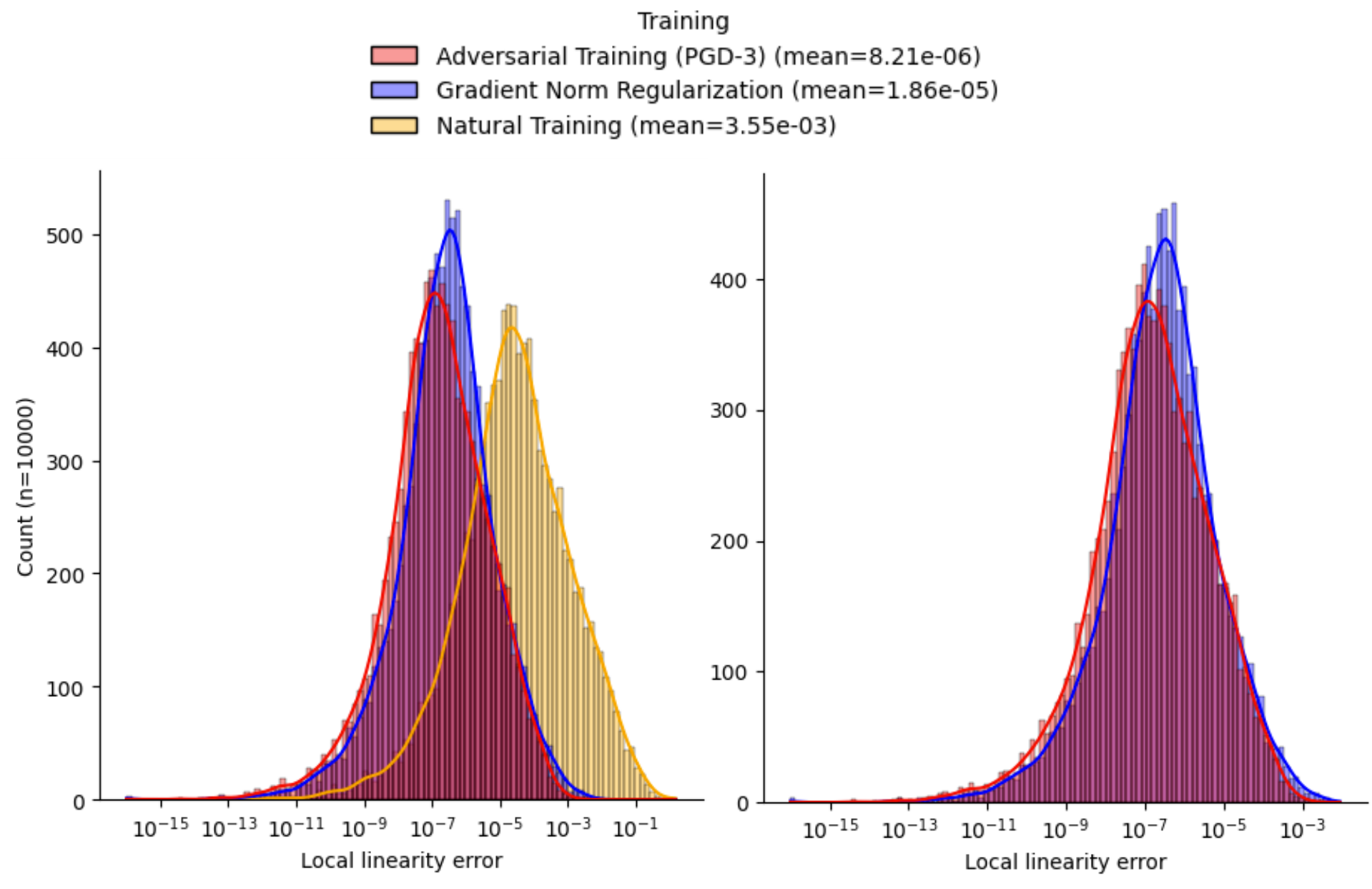}
    \caption{Distribution of local linearity error \cite{rocamora_efficient_2023} over 10000 ImageNet validation images, (a) plots all three models and (b) plots just the robust models. Due to the high variation in scale over images, the graph is plotted with the x-axis in log scale. In (a) we see robust training (both gradient norm regularization and adversarial training) leads to a visible leftward shift of the error distribution by about two orders of magnitude \wrt natural training. Within the robust models, adversarial training has the lowest local linearity error. The error distribution of adversarial training is shifted slightly to the left and has a much lower peak, meaning that it has a greater number of images with very low local linearity error. Architecture is Swin Transformer.}
    \label{fig:linerr}
\end{figure}

\subsection{Loss, $L_1$ Gradient Norm, and Normalized Curvature Along Attack Direction}
\label{sec:attackdir}

\begin{figure}[t]
    \centering
    \includegraphics[width=\linewidth,height=\textheight,keepaspectratio]{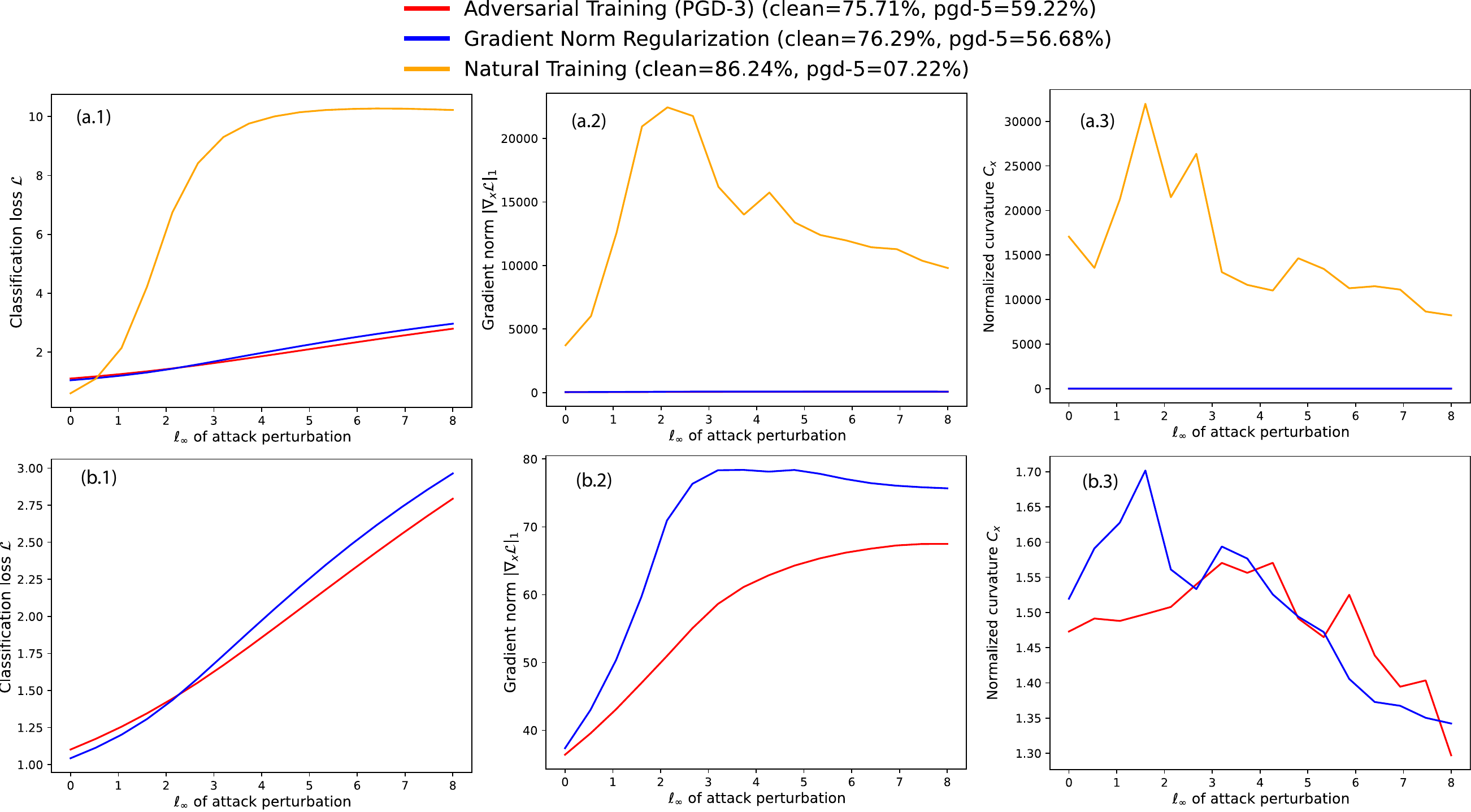}
    \caption{Average loss (left), $L_1$ gradient norm (middle), and normalized curvature (right) \cite{srinivas_efficient_nodate} along the PGD-5 ($\epsilon=4$) attack direction. The top row shows all three models, where we see the extreme brittleness of natural training. The bottom row zooms into the two robust models. Despite their robustness, loss and gradient norm significantly increase along the attack direction for both models. Comparing the two, $L_1$ gradient norm and curvature at the origin are similar, but gradient norm regularization has significantly higher curvature slightly away from the origin, resulting in higher losses and gradient norms along the attack direction. Statistics calculated on 1000 ImageNet validation examples. Power iteration for the normalized curvature done with 2 iterations and initialized from gradients in order to reduce computational cost. Architecture is Swin Transformer.}
    \label{fig:interpolation}
\end{figure}

In this section, we measure how statistics change along the attack direction by linearly interpolating between a clean and attacked input. While we use the relatively weak PGD-5 attack to reduce computational expense, as we can see in the legend of \cref{fig:interpolation} that it is enough to clearly separate the three models in terms of robustness. The interpolated attack is calculated as follows
\begin{equation}
    x(\epsilon) = x + \frac{\epsilon}{4}(\text{PGD-5}_{\epsilon=4}(x) - x)
\end{equation}

\cref{fig:interpolation}.a1 plots average classification loss as function of interpolation $\epsilon$ for the three models. As we can see, the loss of the naturally trained model quickly grows to worse-than-random ($6.9$). \cref{fig:interpolation}.a2 zooms into the two robust models, where we observe two key things: (1) adversarial training has higher loss very close to the origin but lower loss away, and (2) despite their robustness loss still significantly increases as we move in the attack direction.

\cref{fig:interpolation}.b1 and \cref{fig:interpolation}.b2 plots $L_1$ loss-input gradient norm as a function of interpolation $\epsilon$ in the same manner, with similar conclusions. Despite having very similar gradient norms at the origin, gradient norm regularization quickly becomes worse than adversarial training away from the origin. Moreover, both models display a high increase in gradient norm along the attack direction \wrt the value at the origin.

\cref{fig:interpolation}.c1 and \cref{fig:interpolation}.c2 plot normalized curvature. In this particular sample of 1000 images (and using less iterations when calculating hessian spectral norm), normalized curvature at the origin is slightly higher for gradient norm regularization, though they are still quite similar. However, as we step away from the origin in the attack direction, curvature for gradient norm regularization spikes relative to adversarial training. This is consistent with the relative increase in gradient norm \wrt the adversarially trained model seen in \cref{fig:interpolation}.b2.

\FloatBarrier

\begin{figure}[t]
    \centering
  \vspace{0pt}
     \includegraphics[width=0.99\linewidth]{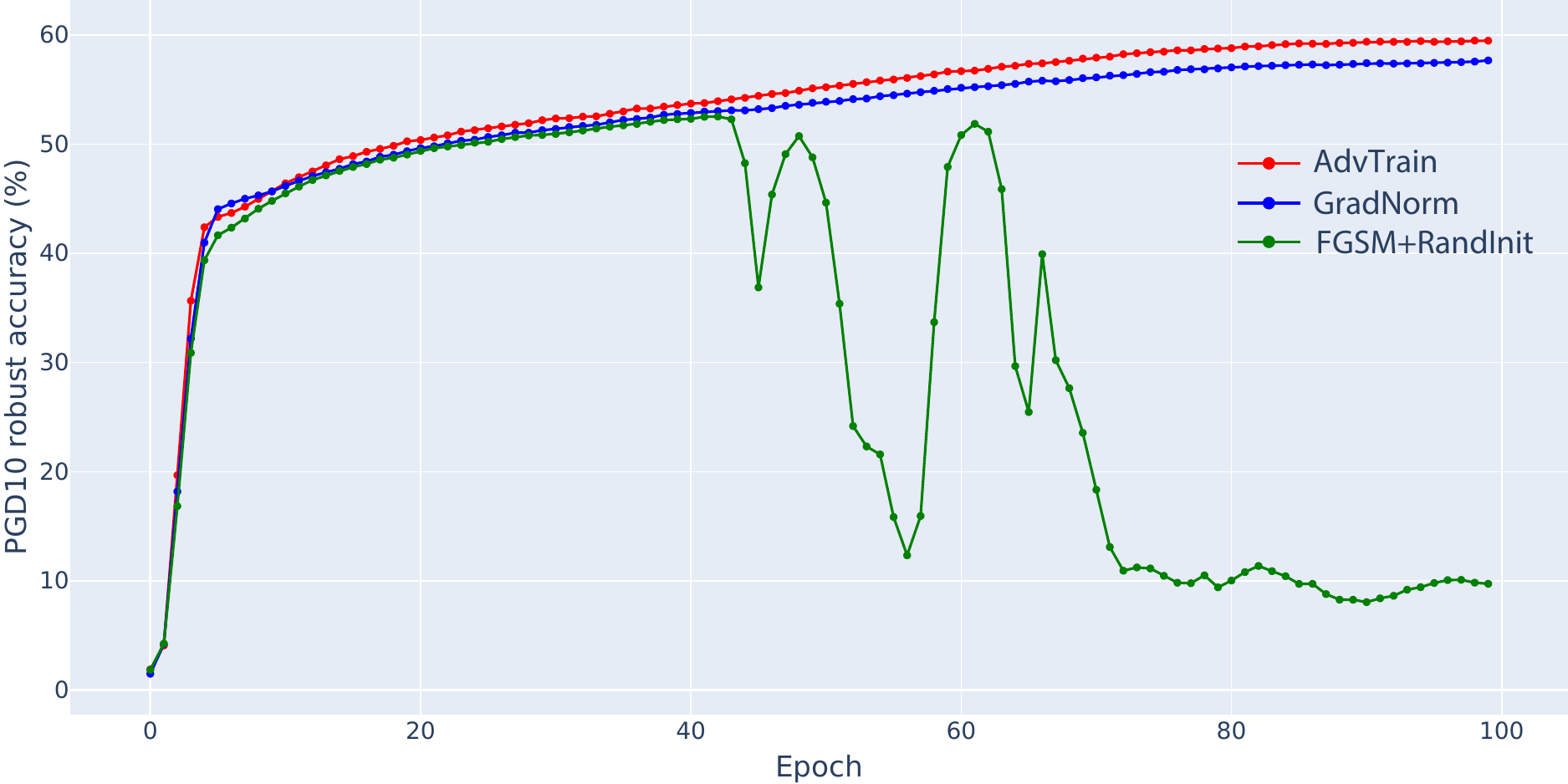}
    \caption{PGD10 ($\epsilon=\frac{4}{255})$ accuracy comparison with FGSM training \cite{wong_fast_2020} for Swin Transformer over 100 epochs.
    FSGM training catastrophically overfits in longer training, as also noted in \cite{wong_fast_2020}. Just before FGSM training's collapse (epoch 43), FGSM is also the worst among the three (0.8\% less robust than GradNorm on PGD10).}
\label{fig:wong}
\end{figure}
\section[B]{Comparison with FGSM Training}
\label{sec:fgsm-comparison}

\cref{fig:wong} plots PGD10 accuracy vs epoch for Adversarial, GradNorm, and FGSM training with random inititialization \cite{wong_fast_2020} for Swin Transformer over 100 epochs. Consistent with Sec. 5.4 of \cite{wong_fast_2020}, we also find FGSM training to catastrophically overfit in longer training, and fail to reach a comparable performance. In other words, FGSM-trained models become robust to FSGM attacks but vulnerable to stronger ones (e.g., at epoch 57: FGSM acc.~52\%, PGD100 acc.~9.78\%). In contrast, GradNorm training does not suffer from such overfitting (\cref{fig:wong}). Just before FGSM training's collapse (epoch 43), FGSM is also the worst among the three (0.8\% less robust than GradNorm on PGD10).

\section[C]{Sanity Checks For Empirical Evaluation of Robustness}
\label{sec:robustness-checklist}

In addition to using the community standard AutoAttack \cite{croce_reliable_2020} for our main robustness evaluations, we performed the suggested checks for rigorous evaluations presented in \cite{carlini_evaluating_2019}, and some others, which we present here.

\subsection{Threat Model}

As stated in \cref{sec:experimental-settings}, the attacker has access to the model end-to-end, and operates under an $L_{\infty}$ constraint of $\frac{4}{255}$, the standard used by the RobustBench public benchmark \cite{croce_robustbench_2020} and prior works \cite{liu_comprehensive_2023}.

\subsection{Adaptive Attacks}

AutoAttack \cite{croce_reliable_2020} is the currently known strongest $L_\infty$ attack, and is the standard in adversarial robustness. Gradient norm regularization minimizes the norm of the cross entropy loss gradient (CE), and AutoAttack uses targeted attacks (FAB-T, APGD-T), boundary attacks (FAB-T), and black-box/gradient-free attacks (Square). 

Moreover, the "plus" version of AutoAttack adds attacks with alternative losses (APGD-DLR) and the un-targeted boundary attack FAB, and only increased attack success by 0.08\%. The latter took over 5x the computational cost to evaluate than the standard AutoAttack, so we only evaluated it for the main GradNorm model of \cref{tab:grad-norm-autoattack-table}.

\subsection{Release Models and Source Code}

These can be found on our \href{https://github.com/adriarm/robustness_input_gradients}{github}.

\subsection{Clean Model Accuracy}

Found in all tables and figures such as \cref{tab:grad-norm-autoattack-table,fig:grad-norm-pgd100-curve,fig:pareto}.

\subsection{Accuracy versus Number of Steps and Perturbation Strength}

\begin{figure}[t]
    \centering
    \includegraphics[width=\linewidth]{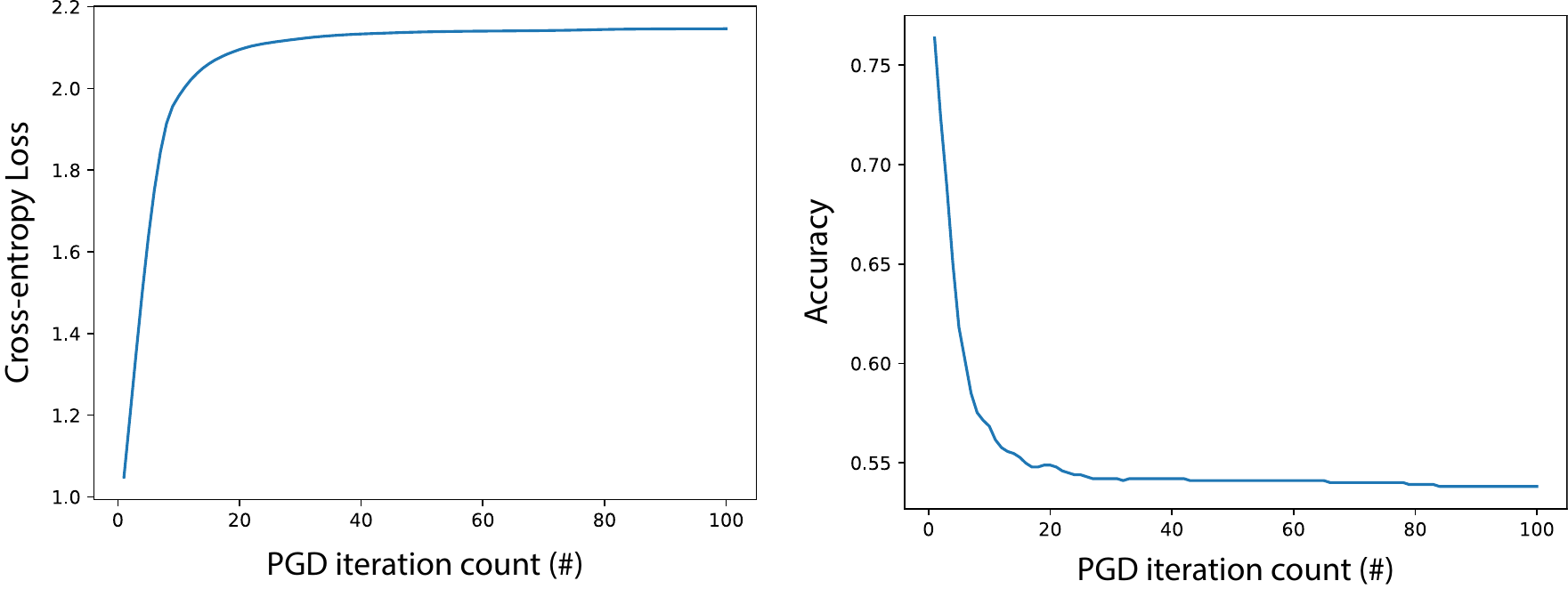}
    \caption{Cross-entropy loss and accuracy as function of PGD iterations ($L_\infty, \epsilon=\frac{4}{255}, \text{step size} \frac{1}{255}$).}
    \label{fig:convergence-sanity}
\end{figure}

See \cref{fig:grad-norm-pgd100-curve} for accuracy vs perturbation strength, where we confirm that accuracy smoothly reaches 0\% as perturbation strength is increased. The following \cref{fig:convergence-sanity} shows smooth convergence of accuracy and loss as the number of PGD($L_\infty, \epsilon=\frac{4}{255}, \text{step size}=\frac{1}{255}$) steps is increased.

\subsection{Transfer Attack from Adversarially Trained Model}

We computed a transfer attack using PGD100($L_\infty, \frac{4}{255}$) as the attack algorithm, the adversarially trained model of \cref{tab:grad-norm-autoattack-table} as the source model, and the Grad. Norm model of \cref{tab:grad-norm-autoattack-table} as the target model. The accuracy obtained was 63.59\%, significantly higher than the 54\% obtained against the same PGD100 attack but using the Grad.Norm model as a source (\ie without transfer). 54\% is itself higher than the accuracy achieved by the model against AutoAttack (51.58\%), also as expected.

\subsection{Convergence of Attacks}

Doubling the iterations of all AutoAttack attacks from 100 to 200 (with the exception of Square, which already uses 5000 queries, for computational reasons), only increased the success rate by 0.02\%, showing convergence of the attack suite.

Additionally, the full AutoAttack, which adds attacks with alternative losses (APGD-DLR) and an untargeted boundary attack (FAB), only increased attack success by 0.08\%.

\subsection{L2 Direction with $L_\infty$ Constraint}

We also performed a sanity check whereby we used the $L_2$ descent direction for PGD100, but keeping the $L_\infty$ constraint, by forgoing the sign operation  and scaling appropriately. The accuracy obtained by the model against this attack was 65\%, higher than that obtained against PGD100 using the $L_\infty$ direction (54\%) as expected.

\begin{table}[t]
    \caption{Results with no noise, gaussian noise ($\sigma=\frac{4}{255}$, 10 epoch finetuned from noiseless GradNorm), and adversarial noise (PGD3, $\epsilon=\frac{4}{255}$, 1 epoch finetune from AdvTrain).}
  \label{tab:non-natural-input}
  \centering
  \resizebox{.90\linewidth}{!}{
  \begin{tabular}{cc|cc}
    \multicolumn{2}{c|}{\bf Method} & \multicolumn{2}{c}{\bf Accuracy} \\
     \cmidrule(lr){1-2}\cmidrule(lr){3-4}
     Training & Noise & Standard & AutoAttack-$L_\infty$\\
    \midrule
    \multirow{3}{*}{Grad. Norm ($\lambda_{\text{CE}}=0.8,\lambda_{\text{GN}}=1.2$)} & None & 77.76 & 51.66 \\
     & Gaussian & \textbf{77.84} & 52.04 \\
     & Adversarial & 76.62 & \textbf{56.49} \\
    \cdashlinelr{1-4}
    Adv. Train. (PGD-3, $\epsilon=4$) & - & 77.20 & 56.12 \\
  \bottomrule
  \end{tabular}}
\end{table}
\section[D]{Gradient Norm Regularization on Non-Natural Inputs}
\label{sec:non-natural-inputs}

While the focus of our work is on robustness using naturaly inputs only, we also investigate the effect of using non-natural inputs for completeness. \cref{tab:non-natural-input} shows regularizing adversarial input slightly improves over AdvTrain; and regularizing random noise input slightly improves over noiseless GradNorm.


\begin{figure}[t]
    \centering
    \includegraphics[width=\linewidth]{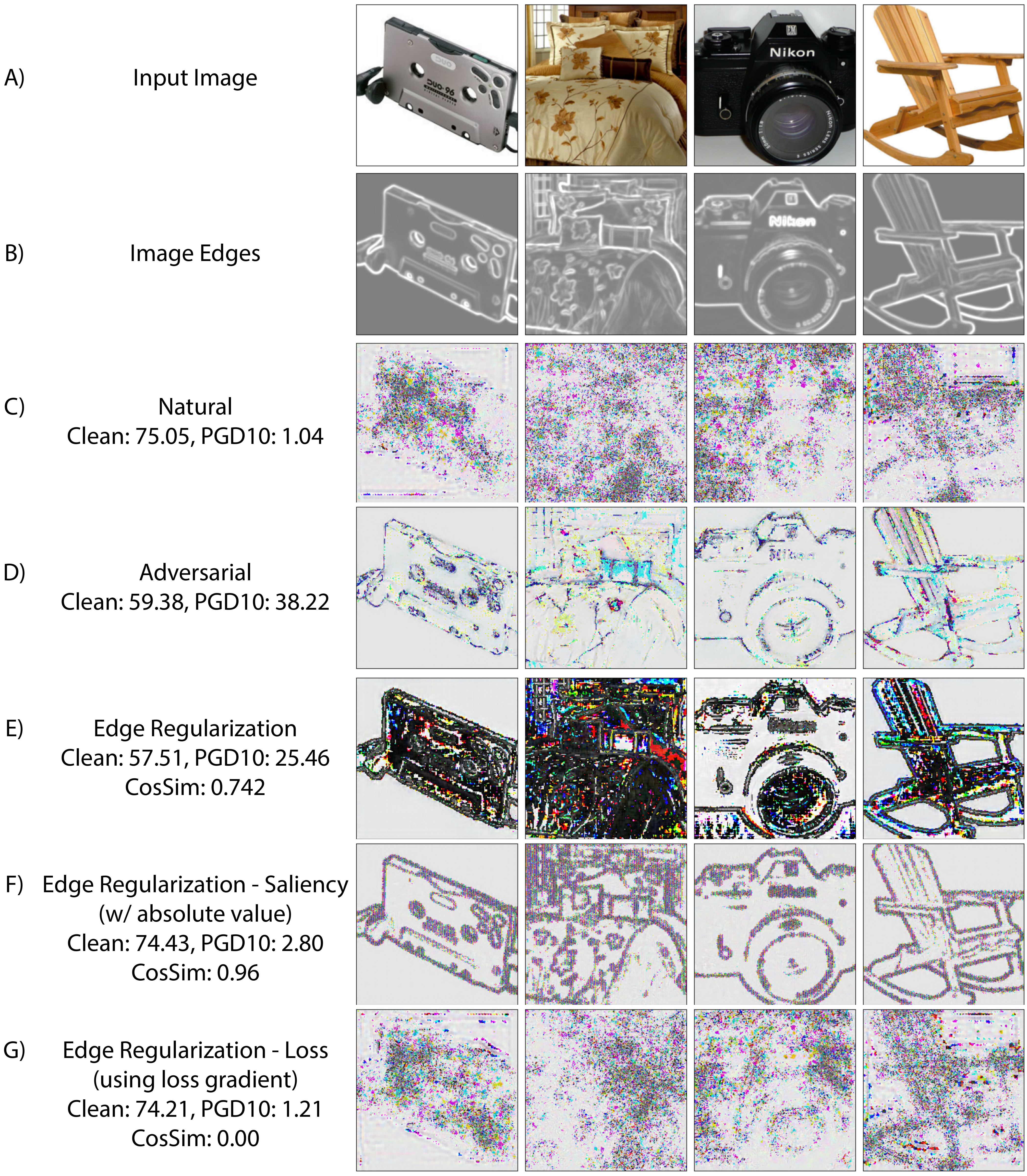}
    \caption{Input image, image edges, and target-class logit \wrt input gradients of ResNet50+GeLU trained with different regularizations.
    Adversarial training (row D) has both image-like and smooth gradients. Edge Regularization (row E) makes gradients image-like and smooth, and non-trivial robustness is obtained. Regularizing saliency maps (row F) leads to a shortcut solution that, despite being image-like, is not smooth due to high frequency sign-flops, and robustness is not obtained. Lastly, regularizing loss gradients (row G), does not optimize, and has non-image-like and non-smooth gradients akin to the naturally trained model (row C). PGD10 attack performed with $\epsilon=\frac{4}{255}$. Gradients normalized to [0, 1] then shifted by 0.4 for visualization purposes only.} 
    \label{fig:gradient-edges-appendix}
\end{figure}
\section[D]{Edge Regularization Alternate Approaches and Training Details}
\label{sec:edge-regularization-appendix}

We first tried to optimize cosine similarity between the loss-input gradient ($\nabla_x \mathcal{L})$) and oriented energy directly. However, optimization did not converge: even past 27 epochs, the cosine similarity loss was still at 0.00. As we can see in \cref{fig:gradient-edges-appendix}.G, gradients become neither image-like nor smooth, and robustness is not achieved.

Next, we attempted to optimize cosine similarity between the saliency map (per-pixel maximum absolute value of the target class input-gradients) \cite{santurkar_image_2019, simonyan_deep_2014} and oriented energy. While the optimization converged in this case, a robust model was not obtained. As we can directly visualize in \cref{fig:gradient-edges-appendix}.E, this is because the usage of the absolute value allowed for a shortcut solution where the gradient sign-flops at high frequency. Robustness is also not achieved.

Instead, we obtained significant results by directly aligning the class gradients ($\nabla_x f_\theta(x)_{y_t}$, \ie, the gradient of target-class logit \wrt input), without the absolute value present in saliency map calculation. The work of \cite{kaur_are_2019} also regularized this quantity in a different fashion, but did not obtain strong results on large benchmarks. With this form of edge regularization, gradients become both image-like and smooth, and a non-trivial robustness of 60\% of SOTA Adversarial Training on AutoAttack is obtained, as can be seen in \cref{fig:gradient-edges-appendix}.D. Convergence takes slightly longer than gradient norm regularization, requiring 128 epochs rather than 100.

\begin{table}[t]
    \caption{Clean and robust accuracies under PGD-50-5 (5 random restarts) and AutoAttack with $\epsilon=\frac{8}{255}$ on Cifar-10 with PRN-18 \cite{he_identity_2016}. Natural training for 50 epochs, Adversarial and GradNorm training for 300.}
  \label{tab:cifar10}
  \centering
  \resizebox{.80\linewidth}{!}{
  \begin{tabular}{@{}lccc@{}}
     & \multicolumn{1}{c}{\bf Clean} & \multicolumn{2}{c}{\bf Robust-($L_\infty,\epsilon=\frac{8}{255}$)}\\
     \cmidrule(lr){2-2} \cmidrule(lr){3-4} 
    {\bf Method} & - & PGD-50-5 & AA \\    
    \midrule
    Natural Training  & 89.38 & 00.05 & 00.00 \\
    Grad. Norm ($\lambda_{\text{CE}}=0.8,\lambda_{\text{GN}}=1.2$) & 77.74 & 46.66 & 39.60 \\
    Adv. Train. (PGD-3, $\epsilon=4$)  & 69.81 & 48.85 & 42.05\\
  \bottomrule
  \end{tabular}}
\end{table}
\section[E]{Cifar-10 Experiments}
\label{sec:cifar10-experiments}

We obtained similar results on CIFAR-10+PreActResnet-18 with $\epsilon=\frac{8}{255}$. As seen in \cref{tab:cifar10}, over 300 epochs AdvTrain (PGD-3) obtained clean acc. 70\%, PGD-50-5 acc. 49\%, and AutoAttack acc. 42\%. GradNorm got clean acc. 78\%, PGD-50-5 acc. 47\%, and AutoAttack acc 39.60\%. The AutoAttack robust accuracy gap is 2.45\% in favor of Adversarial Training, but GradNorm has signficantly higher clean accuracy 78\% vs 70\%.

\section[F]{Training Details}
\label{sec:training-details}

We used the code of Liu \etal \cite{liu_comprehensive_2023} as the basis for our experiments. The training configuration used in our experiments is found below in yaml format.

\subsection{ResNet50}

\subsubsection{Adversarial Training (PGD-3)}

Same as the transformer recipe of Liu \etal \cite{liu_comprehensive_2023}, but shortened to 50 epochs. Training done in evaluation mode.

\begin{verbatim}
# optimizer parameters
opt: adamw
opt_eps: 1.0e-8
opt_betas: null
momentum: 0.9
weight_decay: 0.05
clip_grad: null
clip_mode: norm
layer_decay: null

# lr schedule
epochs: 50
sched: cosine
lrb: 1.25e-3
warmup_lr: 1.0e-6
min_lr: 1.0e-5
epoch_repeats: 0
start_epoch: null
decay_epochs: 15
warmup_epochs: 5
cooldown_epochs: 0
patience_epochs: 0
decay_rate: 0.1

# dataset parameters
batch_size: 256

# augmentation
no_aug: False
color_jitter: 0.4
aa: rand-m9-mstd0.5-inc1
aug_repeats: 0
aug_splits: 0
jsd_loss: False
# random erase
reprob: 0.25
remode: pixel
recount: 1
resplit: False
mixup: 0.8
cutmix: 1.0
cutmix_minmax: null
mixup_prob: 1.0
mixup_switch_prob: 0.5
mixup_mode: batch
mixup_off_epoch: 0
smoothing: 0.1
train_interpolation: bicubic
# drop connection
drop: 0.0
drop_path: 0.0
drop_block: null

# ema
model_ema: True
model_ema_force_cpu: False
model_ema_decay: 0.9998

# adversarial training
attack_eps: 0.01568627450980392 # 4./255.
attack_it: 3
attack_step: 0.01045751633986928 # 8./255./3.
\end{verbatim}

Additionally, the attack step is warmed up linearly over 5 epochs as per Liu \etal \cite{liu_comprehensive_2023}.

\subsubsection{Gradient Norm Regularization}

Same as above, but changing adversarial training for gradient norm regularization with the following weights.

\begin{verbatim}
# gradient norm regularization
ce_weight: 0.5
gradnorm_weight: 0.5
\end{verbatim}

Additionally, \texttt{gradnorm\_weight} is warmed up linearly over 5 epochs.

\subsection{Swin Transformers}

\subsubsection{Adversarial Training (PGD-3)} Exactly the recipe of Liu \etal \cite{liu_comprehensive_2023}.

\begin{verbatim}
# optimizer parameters
opt: adamw
opt_eps: 1.0e-8
opt_betas: null
momentum: 0.9
weight_decay: 0.05
clip_grad: null
clip_mode: norm
layer_decay: null

# lr schedule
epochs: 100
sched: cosine
lrb: 1.25e-3
warmup_lr: 1.0e-6
min_lr: 1.0e-5
epoch_repeats: 0
start_epoch: null
decay_epochs: 30
warmup_epochs: 5
cooldown_epochs: 0
patience_epochs: 0
decay_rate: 0.1

# dataset parameters
batch_size: 512

# augmentation
no_aug: False
color_jitter: 0.4
aa: rand-m9-mstd0.5-inc1
aug_repeats: 0
aug_splits: 0
jsd_loss: False
# random erase
reprob: 0.25
remode: pixel
recount: 1
resplit: False
mixup: 0.8
cutmix: 1.0
cutmix_minmax: null
mixup_prob: 1.0
mixup_switch_prob: 0.5
mixup_mode: batch
mixup_off_epoch: 0
smoothing: 0.1
train_interpolation: bicubic
# drop connection
drop: 0.0
drop_path: 0.0
drop_block: null

# ema
model_ema: True
model_ema_force_cpu: False
model_ema_decay: 0.9998
\end{verbatim}

Additionally, the attack step is warmed up linearly over 5 epochs as per Liu \etal \cite{liu_comprehensive_2023}.

\subsubsection{Gradient Norm Regularization}

Same as above, except for changing adversarial
training for gradient norm regularization (see \cref{eq:gradnorm-objective}) with weights $(\lambda_{\text{CE}}=0.8,\lambda_{\text{GN}}=1.2)$, a halving of the learning rate to $\num{0.625e-3}$, an increase in the batch size from 512 to 532, and an increase in the warm-up learning rate from $\num{1e-6}$ to $\num{1e-5}$.

\begin{verbatim}
# lr schedule
lrb: 0.625e-3
warmup_lr: 1.0e-5

# dataset parameters
batch_size: 532

# gradient norm regularization
ce_weight: 0.8
gradnorm_weight: 1.2
\end{verbatim}

Additionally, \texttt{gradnorm\_weight} is warmed up linearly over 5 epochs.

\end{document}